\pdfoutput=1

\documentclass[11pt]{article}
 
\usepackage[final]{acl}
\usepackage[symbol]{footmisc}

\usepackage{times}
\usepackage{latexsym}
\usepackage{amsmath}
\usepackage{multirow}
\usepackage{booktabs,siunitx}
\usepackage{amssymb}
\usepackage[T1]{fontenc}

\usepackage[utf8]{inputenc}

\usepackage{microtype}

\usepackage{inconsolata}

\usepackage{graphicx}
\graphicspath{{./images/}}

\DeclareMathOperator{\E}{\mathbb{E}}

\newcommand{\method}{LLM-SSC}
\newcommand{\dataset}{\textsc{biorc800}}
\title{Multi-label Sequential Sentence Classification via Large Language Model}

\author{Mengfei Lan,  \hfill Lecheng Zheng, \hfill Shufan Ming, \hfill Halil Kilicoglu\thanks{Corresponding author}
\\
School of Information Sciences\\
  University of Illinois Urbana-Champaign \\
  \texttt{\{mlan3,  lecheng4,  shufanm2,  halil\}@illinois.edu} \\
 }

\begin{document}
\maketitle
\begin{abstract}

Sequential sentence classification (SSC) in scientific publications is crucial for supporting downstream tasks such as fine-grained information retrieval and extractive summarization. However, current SSC methods are constrained by model size, sequence length, and single-label setting. To address these limitations, this paper proposes \method, a large language model (LLM)-based framework for both single- and multi-label SSC tasks. Unlike previous approaches that employ small- or medium-sized language models, the proposed framework utilizes LLMs to generate SSC labels through designed prompts, which enhance task understanding by incorporating demonstrations and a query to describe the prediction target. We also present a multi-label contrastive learning loss with auto-weighting scheme, enabling the multi-label classification task. To support our multi-label SSC analysis, we introduce and release a new dataset, \dataset, which mainly contains unstructured abstracts in the biomedical domain with manual annotations. Experiments demonstrate \method's strong performance in SSC under both in-context learning and task-specific tuning settings. We release~\textsc{biorc800} and our code at: https://github.com/ScienceNLP-Lab/LLM-SSC.

\end{abstract}

\section{Introduction}

With the increasing number of published scientific papers today, researchers face significant challenges in quickly pinpointing needed information. To address this problem, organizing complex paper content according to the rhetorical roles of each sentence in a structured format has garnered interest~\citep{teufel1998sentence, pradhan2003semantic, ruch2007using, lan2024automatic}. Since the rhetorical roles of each sentence are often informed by the context from neighboring sentences (e.g. appendix ~\ref{contextual_dependence}), this task is referred to as sequential sentence classification (SSC)~\cite{cohan-etal-2019-pretrained}. SSC can enable the fine-grained information retrieval~\cite{yepes13}, enhance extractive summarization~\cite{agarwal09}, and improve other downstream tasks. For example, labeling objective sentences in scientific abstracts can support information retrieval based solely on the papers’ objectives.

Existing studies have explored SSC using contextualized language representations. Artificial neural network (ANN)-based SSC methods typically follow a hierarchical structure: an encoding layer to represent word tokens and embed sentences, a context interaction layer to enhance sentence embedding with surrounding context, and a labeling optimization layer to produce optimized sequential labels~\cite{agibetov2018fast, jin18, jiang2019hierarchical, gonccalves2020deep, yamada2020sequential, shang2021span, li21, brack2022cross}. Other research utilizes the masked token objective or transformers, introducing special tokens to encode contextual information and using these tokens to predict sequential labels~\cite{cohan-etal-2019-pretrained}.

Despite promising progress in the SSC task, several gaps remain, including pretrained model size, input sequence length, multi-label annotation, and dataset creation. Specifically, current ANN- and transformer-based methods have only employed moderately sized pretrained models (e.g., word2vec~\cite{jin18}, SciBERT~\cite{cohan-etal-2019-pretrained, li21}), while the application of large language models (LLMs) in SSC is underexplored. Furthermore, the existing transformer-based methods rely on BERT, which is constrained by 512-token sequence length limit~\cite{cohan-etal-2019-pretrained}. Additionally, SSC has not expanded from single-label to multi-label, which is essential since a sentence can serve multiple rhetorical roles within a context~\cite{molla2022overview}. Moreover, the widely used SSC dataset in the biomedical field, PubMed 200K RCT, is automatically generated from structured abstracts in PubMed~\cite{dernoncourt2017pubmed}. However, this dataset does not include unstructured abstracts with free-form writing styles, which may deviate from the common patterns found in structured abstracts~\cite{cohan-etal-2019-pretrained, gonccalves2020deep}.

To bridge these gaps, this paper explores the application of LLMs in multi-label SSC using manually created datasets. We propose \method, a novel unified framework for in-context learning and parameter-efficient finetuning (PEFT) using Gemma-2b~\cite{team2024gemma} for this task. Unlike existing approaches that create contextual representations of sequential sentences, \method\ leverages LLMs to generate SSC labeling results based on designed prompts, which include a demonstration part to showcase the task and a query part to introduce the prediction target. To address the challenge of multi-label annotation, we design an auto-weighting multi-label contrastive learning loss that relaxes the constraint of formation of positive and negative pairs in the contrastive learning and reweights the importance of positive and negative pairs based on their label information. 

Our contributions are as follows:
\begin{itemize}
\item We present \method, the first LLM-based framework supporting both single- and multi-label SSC that integrates complete contextual information within the prompt and consider neighboring context when making predictions. 
\item We propose a novel multi-label contrastive learning loss with auto-weighting scheme to reweight the importance of negative pairs.
\item We introduce and release \textsc{biorc800}, a manually annotated multi-label SSC dataset mainly using unstructured abstracts from the biomedical field using rhetorical labels (Background, Objective, Methods, Results,  Conclusions, and Other).
\item Extensive experiments demonstrate the strong capability of \method\ in SSC tasks under both in-context learning and parameter-efficient finetuning settings.
\end{itemize}


\section{Methods}
\begin{figure*}[t] 
\includegraphics[width=\textwidth]{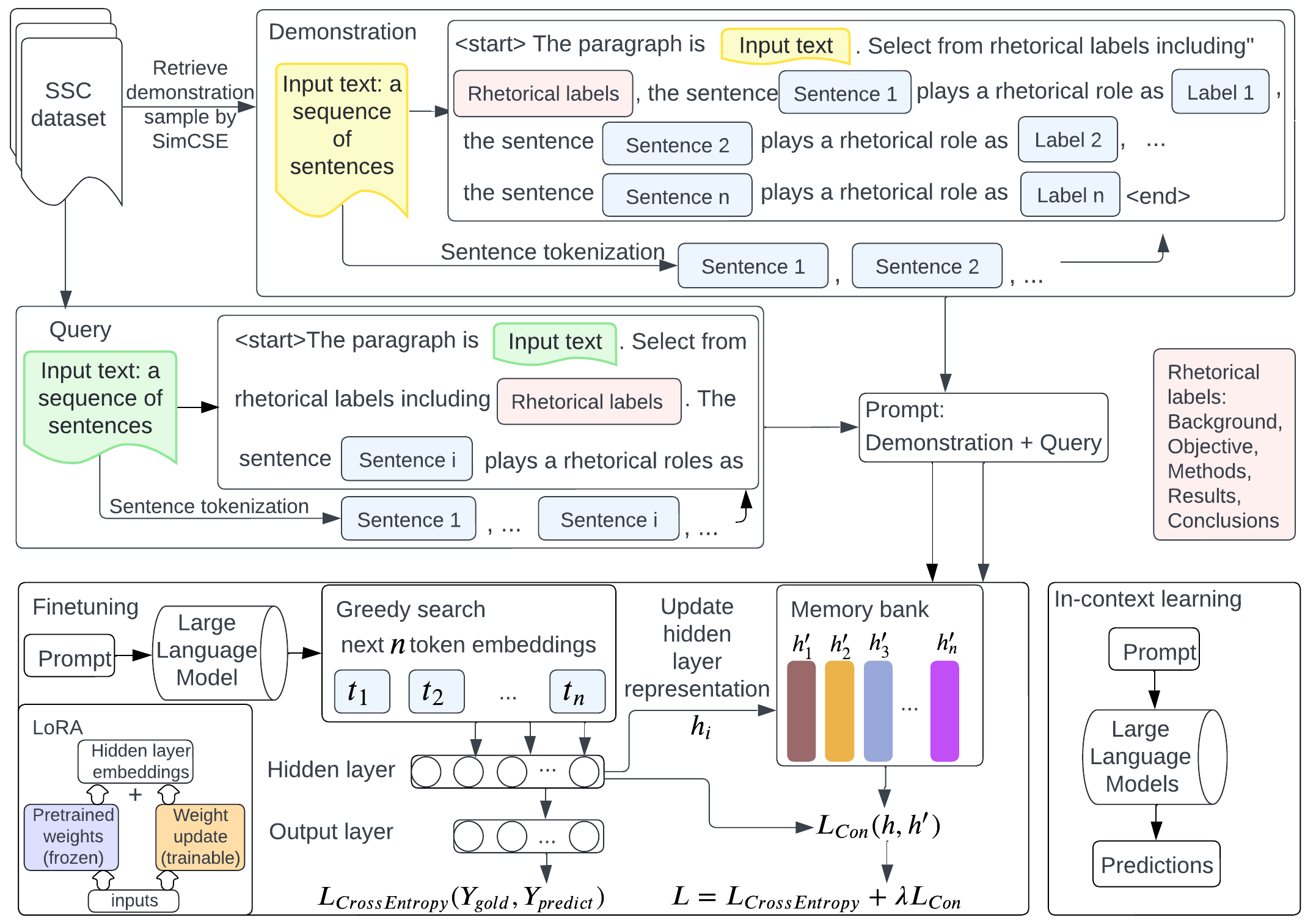}
\centering
\caption{Structure of our LLM-based in-context learning and finetuning for SSC. }
\label{method}
\centering
\end{figure*}

In this section, we first introduce the notation and then present \method, an LLM-based framework for sequential sentences in-context learning and parameter-efficient finetuning, integrating complete contextual information within the prompt and consider neighboring context when making predictions. To enable the multi-label classification, we propose auto-weighting multi-label contrastive learning loss. The overview of the proposed framework is shown in Figure~\ref{method}.   

\subsection{Notation}

We approach SSC as a task of conditional text generation. Specifically, for an SSC dataset with $S$ text sequences, we denote $S_i$ as the $i^{th}$ text sequence, $S_{ij}$ as the $j^{th}$ sentence in $S_i$, $C_{i}$ as the context where the sentence is located ($C_{i} = concate(S_{i1}, S_{i2}, ... S_{in}$)), and $Y_{ij}$ as the SSC label of $S_{ij}$. 
Our goal is to model the probability of generating the SSC label $Y_{ij}$. 

\subsection{In-context Learning}
\label{in_context_learning}
We utilize in-context learning to leverage the power of LLMs for this task. An overview of the ICL framework is provided in Figure \ref{method}. A prompt is created by combining a demonstration context with a query, which is then fed into the language model to generate the prediction. The demonstration samples are selected from the training set based on cosine similarity scores between the training samples and the prediction target sentence. These similarity scores are calculated using embeddings generated by the SimCSE pre-trained model~\cite{gao-etal-2021-simcse}\footnote{https://huggingface.co/princeton-nlp/sup-simcse-roberta-large}. Given a demonstration sample $D_i$, the label $Y_i$ for the $i^{th}$ sentence in $D_i$, and the set of rhetorical label candidates $U$, the demonstration part of the prompt $D_{prompt}$ is constructed as:

\textit{<Start> The paragraph is [$D_i$]. Select from rhetorical labels including  [$U$], the sentence [$D_{i1}$] plays a rhetorical role as <[$Y_{i1}$]>, the sentence [$D_{i2}$] plays a rhetorical role as <[$Y_{i2}$]>, ..., the sentence [$D_{in}$] plays a rhetorical role as <[$Y_{in}$]> <End>}. 

Then we create the query part of prompt. Given the prediction target sentence $S_{ij}$ and the context $C_i$ where the target sentence is located, the query portion of the prompt $Q_{prompt}$ is formatted as: 

\textit{<Start> The paragraph is [$C_i$]. Select from rhetorical labels including [$U$], the sentence [$S_{ij}$] plays a rhetorical role as} 

The input prompt is constructed by combining the demonstration $D_{prompt}$ and query $Q_{prompt}$:

\begin{align}
     X_{ICL} = D_{prompt}||Q_{prompt}
\end{align}

The goal for in-context learning is to generate the SSC label $Y_{predict}$ given the input prompt $X_{ICL}$:
\begin{align}
Y_{predict} =  \arg\max_{Y}P(Y|X_{ICL}) 
\end{align}
where $Y_{predict}$ denotes the generated label that maximizes the conditional probability given the input prompt $X_{ICL}$. $P(Y|X_{ICL})$ denotes the conditional probability of generating outcome $Y$ given the prompt $X_{ICL}$. Some actual prompts are provided in Appendix~\ref{in_context_learning_prompts}.

\subsection{Task-specific Model Tuning}

Although LLMs can recognize SSC labels using ICL due to their generalization ability without any parameter tuning, ICL underperforms the fine-tuning methods in text classification tasks~\cite{dong2022survey, sun-etal-2023-text, wadhwa2023revisiting}. To further explore the LLM application in SSC, we design a parameter-efficient fine-tuning framework of LLM. Figure~\ref{method} presents an overview of the fine-tuning framework. 

\textbf{Supervision with Demonstration}. To bridge the gap between the pretrained model's original objective of predicting the next token and the goal of SSC to have the model generate the specific label for the classification target, we include one SSC demonstration within the input to guide the model's response. The format of the demonstrations and queries used in fine-tuning prompts $X_{finetune}$ is the same as that used in ICL (as described in Subsection \ref{in_context_learning}). The tuning process modifies how the model adjusts the given demonstration and query within the prompt to generate appropriate token sequence $\hat{t}$:
\begin{align}
    \label{generation_function}
    \hat{t} = \arg\max_{t}P(t|X_{finetune})
    \\
    \hat{t} = \{t_1, t_2, ..., t_i, ...\}
\end{align}
where $t_i$ denotes the hidden state of the $i$th token in the generated sequence. 

\textbf{Think Before Speak}. Previous research found that giving space for the LLM model to produce additional tokens (delays) before generating the expected answer shows performance gains across various downstream tasks~\cite{goyal2023think}. In our preliminary analysis, we observe a similar phenomenon. When employing the ICL approach outlined in Subsection \ref{in_context_learning}, the model does not immediately generate the expected SSC label but first produces tokens not present in the label set. Motivated by this finding, we design the space-thinking mechanism~\cite{goyal2023think} to provide some space for LLM to think before generating the expected answer.

The space-thinking mechanism requires LLM to generate the next $n$ tokens after the prompt using greedy search. We assume that the predicted results are contained within one or more of these generated tokens. Therefore, we create a verbalizer to map the multiple generated tokens to the label space by concatenating the hidden states from the last layer of the generated tokens and feeding the combined results into a two-layer MLP. Specifically, given the prompt $X_{finetune}$, the next $n$ token hidden states after the input context are generated as in equation~\ref{generation_function} and concatenated:
\begin{align}
\label{hidden_states_function}
    e_i = concat(t_1, t_2, ..., t_n)
\end{align}

Then a two-layer MLP is applied to map the concatenated representation to the label space: 
\begin{align}
    h_i &= ReLU(w_1e_i + b_1) \\
    p_{i,predict} &= \sigma(w_2h_i + b_2)
\end{align}
We use the cross entropy loss to compare the difference between the prediction probability $p_{i, predict}$ and the golden standard $y_{i, gold}$ for the $i$-th sequence, where $N$ denotes the number of classes: 
\begin{align}
L_{CrossEntropy} = -\sum\limits_{i=1}^{N}y_{i, gold}\log(p_{i, predict})
\end{align}
 
\textbf{Parameter-efficient Fine-tuning}. Instead of fine-tuning all model parameters, we leverage low-rank adaptation (LoRA) method to tune the LLM in a parameter-efficient way~\cite{hu2021lora}. Lora keeps the pre-trained weights frozen and introduces trainable low-rank matrices in each layer of the transformer to approximate the weight updates needed for fine-tuning. It helps to reduce the computational cost and increase the memory efficiency during the tuning process.

\subsection{Auto-weighting Multi-label Contrastive Learning}

Supervised Contrastive Learning~\cite{DBLP:conf/nips/KhoslaTWSTIMLK20} has been widely employed in fine-tuning language models~\cite{DBLP:conf/iclr/GunelDCS21, chen2022dual, xie2022multi}. These methods typically construct positive and negative pairs based on the equivalence of label vectors in multi-class classification problems~\cite{DBLP:conf/sdm/ZhengZH23}. However, in the multi-label setting, treating two sentences with the same label vector as a positive pair is impractical due to the exponential growth in the number of unique label vectors with more labels ($2^m$ unique label vectors for $m$ binary labels), resulting in a scarcity of positive pairs for sentences with rare label vectors. In the worst-case scenario, it may be impossible to find two sentences with identical label vectors, thereby hindering the formation of positive pairs for contrastive learning. Additionally, minimizing the similarity of negative pairs in contrastive learning introduces the class collision issue~\cite{zheng2021weakly, zheng2022contrastive, DBLP:conf/kdd/ZhengJLTH24}, where sentences with similar label vectors are erroneously pushed apart in the latent space, leading to sub-optimal solutions.

To address these issues, we propose an auto-weighting multi-label contrastive learning loss (WeighCon). Instead of requiring identical label vectors for positive pairs, we relax this constraint by forming positive pairs if two sentences share at least one common positive class. We introduce an auto-weighting scheme and propose the following multi-label contrastive learning loss:
\begin{align}
\label{con_loss}
L_{con} &= - \sum_{c=1}^m \E_i\E_{j\in P_i(c)}\frac{\alpha_{ij}\text{sim}(h_i, h_j)}{\sum_{k}(1 - \alpha_{ik})\text{sim}(h_i, h_k)} \\
\nonumber \alpha_{ij} &= \sigma(MLP(y_i, y_j))
\end{align}
Here, $\text{sim}(h_i, h_j)=\exp(\frac{h_i h_j^T}{|h_i||h_j|})$ is the exponential of the cosine similarity measurement, $\sigma(\cdot)$ is the sigmoid function, $m$ is the number of labels, and $P_i(c)=\{j|y_i(c)=y_j(c)=1\}$ represents the set of sentences with the same $c^{th}$ label as the $i^{th}$ sentence. The weighting function $\alpha_{ij}$, parameterized by a one-layer MLP, takes two label vectors as input and outputs a scalar indicating the similarity between sentence representations $h_i$ and $h_j$. Intuitively, $\alpha_{ij}$ is large when $y_i$ and $y_j$ are similar, peaking when they are identical. To mitigate the class collision issue, we use $1 - \alpha_{ik}$ to reweight the importance of negative pairs in the denominator of Eq.~\ref{con_loss}. The weight (i.e., $1 - \alpha_{ik}$) of a negative pair is large when label vectors differ significantly, decreasing as label vectors become more similar. By reducing the weights of negative pairs with similar label vector, we mitigate the negative impact of these negative pairs in minimizing the proposed contrastive loss. Furthermore, given the complexity and large-scale parameters of the LLM, we incorporate supervised contrastive learning supported by a memory bank~\cite{he2020momentum} into the training objective, thus reducing the memory requirement. The final loss function for task-specific model tuning contains two items and $L_{Con}$ is weighted by a scaling factor $\lambda$ (default 0.1):

\begin{align}
    L = L_{CrossEntropy} + \lambda L_{Con}
\end{align}

\section{Experiments}
In this section, we first present the a new dataset named \textsc{biorc800}, a manually annotated multi-label SSC dataset mainly using unstructured abstracts from the biomedical field. Then, we evaluate the effectiveness of our proposed \method\ by comparing it with state-of-the-art SSC methods and other contrastive learning based regularization. Additionally, we experiment on in-context learning setting and conduct an ablation study to further validate the assumptions outlined in the previous sections.

\begin{table}[]
\resizebox{\columnwidth}{!}{%
\begin{tabular}{lll}
\hline
\multicolumn{3}{c}{\textsc{biorc800} Statistics}                                            \\ \hline
\multicolumn{1}{l|}{Overall}          & 7911 (Sentences) & 800 (Docs)        \\ \hline
\multicolumn{1}{l|}{Multi-label Sentences}  & 452 (Sentences)  & Percentage: 5.7\% \\ \hline
\multicolumn{1}{l|}{Doc Length (sentence)}  & Avg: 9.89        & Std: 2.68         \\ \hline
\multicolumn{1}{l|}{Sentence Length (word)} & Avg: 23.04       & Std: 11.10        \\ \hline
\multicolumn{1}{l|}{Label Distribution}     & BACKGROUND       & 1252, 15.8\%      \\
\multicolumn{1}{l|}{}                       & OBJECTIVE        & 827, 10.5\%       \\
\multicolumn{1}{l|}{}                       & METHODS          & 2319, 29.3\%      \\
\multicolumn{1}{l|}{}                       & RESULTS          & 2757, 34.9\%      \\
\multicolumn{1}{l|}{}                       & CONCLUSIONS      & 1114, 14.1\%      \\
\multicolumn{1}{l|}{}                       & OTHER            & 112, 1.4\%        \\ \hline
\end{tabular}%
}
\caption{\textsc{biorc} Statistics}
\label{biorc800}
\end{table}

\subsection{Datasets}
\paragraph{Multi-label SSC Dataset: \textsc{biorc800} }

To enhance our multi-label sequential sentence classification (SSC) analysis and address the lack of manual SSC labels in unstructured biomedical texts, we manually annotated a corpus comprising 700 unstructured and 100 structured PubMed abstracts. Previous studies show that though sentences of the structured abstracts have author-assigned rhetorical categories, the categories might be erroneous~\cite{cohan-etal-2019-pretrained, gonccalves2020deep} (e.g. appendix~\ref{semantic_coherence}). Therefore, we re-annotated those sentences from 100 structured abstracts to more accurately reflect their category. The collected RCT abstracts were sampled from PubMed Central Open Access subset\footnote{https://www.ncbi.nlm.nih.gov/pmc/tools/openftlist/} using a modified version of Cochrane's sensitivity and precision-maximizing query for RCTs. The annotation utilized the multi-label approach and followed the annotation schema of Background, Objective, Methods, Results, Conclusions, and Other.



We evaluated how consistently pairs of annotators agreed on sentence-level annotations using Cohen's $\kappa$ over several stages. In the first stage, four annotators, who are also experts in biomedical text mining, used the first version of guideline to annotate the same 50 abstracts, with agreement scores between pairs ranging from 0.757 to 0.856. After discussing challenges and updating the guidelines, the second stage involved annotating a new set of 50 abstracts, improving the agreement scores to between 0.784 and 0.879. In the third stage, after further discussions and guideline updates (guideline final version: Appendix~\ref{guideline}), the remaining 700 abstracts were divided equally among the annotators. Finally, all 800 abstracts were combined, and one senior annotator reconciled the final set of labels. Compared to the author-assigned labels for the 100 structured abstracts, our re-annotation changed 4.1\% sentence labels. We finally splited the 800 abstracts into training (480 abstracts), development (160), and test sets (160), keeping the proportion of structured vs. unstructured abstracts the same in all three (12.5\% - 87.5\%). The descriptive statistics of \textsc{biorc800} are shown in Table~\ref{biorc800}. Additional dataset information is provided in Appendix~\ref{detailed_stat_append}. 

\paragraph{Single-label SSC Dataset}
In addition to the proposed \textbf{\textsc{biorc800}} dataset, we test the models on the following two datasets in our experiments: 

\textbf{\textsc{CS-Abstract}}~\cite{gonccalves2020deep} contains 654 abstracts selected from computer science literature classified into Background, Objective, Methods, Results, and Conclusions sentences. It is the most recently published computer science RSC dataset annotated by crowdsourcing and collective intelligence\footnote{https://github.com/sergiog95/csabstracts}.


\textbf{\textsc{PubMed 20K RCT}}~\cite{dernoncourt2017pubmed} contains 20k structured biomedical abstracts of randomized controlled trials with sentences automatically classified based on the author-assigned annotations as background, objective, method, result, or conclusion\footnote{https://github.com/Franck-Dernoncourt/pubmed-rct/tree/master/PubMed\_20k\_RCT}. 

\textbf{\textsc{ART-CoreSC}}~\cite{liakata-etal-2010-corpora} is a multi-domain dataset,  containing sentence-level scientific discourse annotation for 265 full papers selected from physics, chemistry, and biochemistry fields.  In this SSC task, only the abstracts of these papers are used. The sentences in abstracts are annotated as background, hypothesis, motivation, objective, goal, methods, results, observation, experiment, or conclusion\footnote{https://live.european-language-grid.eu/catalogue/corpus/972/download/}.  
\subsection{Baselines}
We selected SSC methods that achieved state-of-the-art performance on SSC datasets and had publicly accessible code as baselines for testing on our \textsc{biorc800} dataset. To adapt these methods, which were originally designed for single-label settings, to multi-label prediction, we modified the code provided by the authors by applying a threshold of 0.4 (chosen empirically to balance precision and recall for each label) to the output label probabilities. 
\paragraph{Hierarchical Sequential Labeling Network} (HSLN)~\cite{jin18} creates bi-RNN sentence representation, followed by attention-based pooling and a bi-LSTM layer to add contextual information from surrounding sentences. Finally, a CRF layer is concatenated to optimize the label sequence \footnote{https://github.com/jind11/HSLN-Joint-Sentence-Classification}.

\paragraph{Sequential Sentence Classification} (SSC)~\cite{cohan-etal-2019-pretrained} employs BERT model~\cite{devlin2019bert} to encode both the semantics of the target sentence and the sequence's contextual information into a [SEP] token appended after the target sentence. This [SEP] token acts as the target sentence's representation, used to predict the rhetorical label\footnote{https://github.com/allenai/sequential\_sentence\_classification}.

\paragraph{Scientific Discourse Tagging} (SDT)~\cite{li21}  uses token embeddings from SciBERT~\cite{beltagy-etal-2019-scibert}, an LSTM layer to encode sentences, and a bi-LSTM layer for sentence labeling, followed by a CRF layer with BIO tagging scheme to optimize the order of sequence labels\footnote{https://github.com/jacklxc/ScientificDiscourseTagging}. 

\paragraph{SciBERT-HSLN}~\cite{brack2022cross} is built upon the HSLN model with SciBERT~\cite{beltagy-etal-2019-scibert} as word embeddings\footnote{https://github.com/arthurbra/sequential-sentence-classification}. 

We also include a contrastive learning baseline:

\paragraph{HeroCon}~\cite{zheng2022contrastive} is designed for multi-view and multi-label learning that applies weight to positive and negative label pairs by hamming distance of two label representations\footnote{https://github.com/Leo02016/HeroCon}.

\subsection{Experimental Setup}
\paragraph{Implementation and Evaluation} We select Gemma-2b~\cite{team2024gemma} as the backbone due to its lightweight design and advanced performance across various natural language tasks. This model supports an input sequence length of up to 8192 tokens, which we adopt as the maximum length for in-context learning. For fine-tuning, however, We limit the sequence length to 1200 tokens, a value chosen empirically to fit within the 40GB RAM of the GPU (experiments are performed on NVIDIA A100) and mitigate excessive computational demands and high memory usage. If the input sequence length exceeds this limit, we remove the demonstration part of the input and only input the query part. To evaluate the proposed in-context learning method, we use the training set samples as demonstrations and test set samples as query. For validating the fine-tuning method, we tune the parameters on the training set, select the best model on the validation set, and finally test and report the performance of the selected model on the test set. We use PEFT~\footnote{https://huggingface.co/docs/peft/en/index} package to tune model using LoRA. Our default model is trained with the AdamW optimizer with zero weight decay. 

\paragraph{Thresholding}
When evaluating the proposed model on the multi-label dataset (\textsc{biorc800}), we apply dynamic thresholding, which utilizes different probability thresholds for each label. The optimal threshold for each label is determined by maximizing the label-specific F$_{1}$ score on the validation set. For single-label datasets, we apply softmax function to select the best label.

\subsection{Results and Discussion}

\subsubsection{In-context Learning}



\begin{table*}[]
\centering
\resizebox{\textwidth}{!}{
\begin{tabular}{c|cc|cc|cc|cc}
\hline
\multirow{2}{*}{Dataset} & \multicolumn{2}{c|}{0-shot} & \multicolumn{2}{c|}{1-shot} & \multicolumn{2}{c|}{5-shot} & \multicolumn{2}{l}{10-shot} \\ \cline{2-9} 
 & \multicolumn{1}{c|}{Micro F1} & Macro F1 & \multicolumn{1}{c|}{Micro F1} & Macro F1 & \multicolumn{1}{c|}{Micro F1} & Macro F1 & \multicolumn{1}{c|}{Micro F1} & Macro F1 \\ \hline
\textsc{biorc800} & \multicolumn{1}{c|}{0.476} & 0.322 & \multicolumn{1}{c|}{0.642} & 0.507 & \multicolumn{1}{c|}{\textbf{0.733}} & \textbf{0.656} & \multicolumn{1}{c|}{0.159} & 0.068 \\ \hline
\textsc{CS-abstract} & \multicolumn{1}{c|}{0.468} & 0.331 & \multicolumn{1}{c|}{0.515} & 0.454 & \multicolumn{1}{c|}{\textbf{0.581}} & \textbf{0.562} & \multicolumn{1}{c|}{0.563} & 0.541 \\ \hline
\textsc{PubMed 20K RCT} & \multicolumn{1}{c|}{0.171} & 0.131 & \multicolumn{1}{c|}{0.642} & 0.546 & \multicolumn{1}{c|}{\textbf{0.712}} & \textbf{0.659} & \multicolumn{1}{c|}{0.579} & 0.528 \\ \hline
\textsc{ART-CoreSC} & \multicolumn{1}{c|}{0.064} & 0.029 & \multicolumn{1}{c|}{0.207} & 0.100 & \multicolumn{1}{c|}{0.193} & 0.103 & \multicolumn{1}{c|}{\textbf{0.217}} & \textbf{0.102} \\ \hline

\end{tabular}}
\caption{In-context learning results with different number of demonstrations (shots). }
\label{tab:in_context_learning}
\end{table*}

In this subsection, we evaluate the model using 0-shot (no demonstrations in the prompt), 1-shot (one demonstration), 5-shot, and 10-shot settings, where the shots are chosen from the training set using SimCSE ranking, and the queries are from the test set. Table~\ref{tab:in_context_learning} presents the performance of our in-context learning approach across all datasets. Specifically, we have the following observations: (1) \method\ with 5-shot setting achieves the highest micro F1 scores for \textsc{biorc800}, \textsc{CS-abstract}, and \textsc{PubMed 20K RCT}; (2) in the zero-shot setting, in-context learning on \textsc{biorc800} and \textsc{CS-abstract} datasets consisting of entire paragraphs of unstructured text achieves micro F$1$ scores as 0.476 and 0.468, and macro F$1$ scores as 0.322 and 0.331. This demonstrates the large language model's generative ability to recognize without seen any training data; (3) the 0-shot in-context learning performance on \textsc{PubMed 20K RCT} is relatively poor, where sentences are organized by the original authors to meet specific structural requirements at the expense of contextual dependence on each other; (4) the performance improvement from 0-shot to 1-shot emphasizes the importance of including samples in the prompt for LLMs to understand SSC tasks; (5) when provided with more samples (10-shots) on \textsc{biorc800}, \textsc{CS-abstract}, and \textsc{PubMed 20K RCT}, the performance are not as good as with fewer examples. We attribute this drop to the additional information introducing bias and confusing the large language model to capture the task-related features from the additional samples; (6) the poor in-context learning results yielded on \textsc{CoreSC} can be attributed to the dataset's fine-grained rhetorical categories, which are challenging for large language models to recognize by simply relying on general common-sense reasoning or surface-level patterns without more detailed guidelines.


\subsubsection{Task-specific Model Tuning}

\begin{table*}[]
\resizebox{\textwidth}{!}{%
\begin{tabular}{cc|cc|cc|cc|cc}
\hline
\multicolumn{2}{c|}{\multirow{2}{*}{}} &
  \multicolumn{2}{c|}{\textsc{biorc800}} &
  \multicolumn{2}{c|}{\textsc{CS-abstract}} &
  \multicolumn{2}{c|}{\textsc{PubMed 20K RCT}} &
  \multicolumn{2}{l}{\textsc{ART-CoreSC}} \\ \cline{3-10} 
\multicolumn{2}{c|}{} &
  \multicolumn{1}{c|}{Micro F1} &
  Macro F1 &
  \multicolumn{1}{c|}{Micro F1} &
  Macro F1 &
  \multicolumn{1}{c|}{Micro F1} &
  Macro F1 &
  \multicolumn{1}{c|}{Micro F1} &
  \multicolumn{1}{c}{Macro F1} \\ \hline
\multicolumn{2}{c|}{HSLN~\cite{jin18}} &
  \multicolumn{1}{c|}{0.849} &
  0.826 &
  \multicolumn{1}{c|}{0.723} &
  0.652 &
  \multicolumn{1}{c|}{0.919} &
  0.870 &
  \multicolumn{1}{c|}{0.400} &
  0.163 \\ \hline
\multicolumn{2}{c|}{SSC~\cite{cohan-etal-2019-pretrained}} &
  \multicolumn{1}{c|}{{\underline{0.905}}} &
  0.892 &
  \multicolumn{1}{c|}{\textbf{0.780}} &
  \underline{0.714} &
  \multicolumn{1}{c|}{0.924} &
  0.859 &
  \multicolumn{1}{c|}{0.470} &
  0.253 \\ \hline
\multicolumn{2}{c|}{SDT~\cite{li21}} &
  \multicolumn{1}{c|}{-} &
  - &
  \multicolumn{1}{c|}{0.767} &
  0.653 &
  \multicolumn{1}{c|}{\textbf{0.940}} &
  \textbf{0.903} &
  \multicolumn{1}{c|}{\textbf{0.534}} &
  { 0.270} \\ \hline
\multicolumn{2}{c|}{\begin{tabular}[c]{@{}c@{}}SciBERT-HSLN\\ ~\cite{brack2022cross}\end{tabular}} &
  \multicolumn{1}{c|}{0.902} &
  0.897 &
  \multicolumn{1}{c|}{0.765} &
  0.712 &
  \multicolumn{1}{c|}{{ \underline{0.931}}} &
  { \underline{0.882}} &
  \multicolumn{1}{c|}{0.467} &
  0.246 \\ \hline
\multicolumn{1}{c|}{\multirow{2}{*}{LLM-SSC}} &
  \begin{tabular}[c]{@{}c@{}}with HeroCon\\ ~\cite{zheng2022contrastive}\end{tabular} &
  \multicolumn{1}{c|}{0.902} &
  {\underline{0.906}} &
  \multicolumn{1}{c|}{0.767} &
  {\underline{0.714}} &
  \multicolumn{1}{c|}{0.921} &
  0.871 &
  \multicolumn{1}{c|}{0.503} &
  0.260 \\ \cline{2-10} 
\multicolumn{1}{c|}{} &
  \begin{tabular}[c]{@{}c@{}}with WeighCon\\ (Ours)\end{tabular} &
  \multicolumn{1}{c|}{\textbf{0.907}} &
  \textbf{0.912} &
  \multicolumn{1}{c|}{{\underline{0.768}}} &
  \textbf{0.716} &
  \multicolumn{1}{c|}{0.925} &
  0.879 &
  \multicolumn{1}{c|}{{\underline{0.524}}} &
  \textbf{0.282} \\ \hline
\end{tabular}%
}
\caption{Task-specific model tuning results. In \method, the next 2 tokens are generated. The SDT model performance on \textsc{biorc800} is not reported since it uses a BIO tagging mechanism to block different rhetorical sections within a paragraph, making it unsuitable for multi-label classification. }
\label{task_specific_model_tuning}
\end{table*}
Table~\ref{task_specific_model_tuning} presents the performance of the task-specific model-tuning methods. Under the multi-label setting, we observe: (1) our \method\ with WeighCon achieves the highest micro and macro F1 scores (0.907 and 0.912, respectively) when tested on the \textsc{biorc800} dataset; (2) compared to HeroCon, the proposed WeighCon yields better performance, demonstrating its effectiveness with the auto-weighting design; (3) the SSC model delivers the second-best micro F1 score (0.905), together with the proposed LLM-SSC, highlighting the effectiveness of transformer-based methods in multi-label SSC; (4) although the SDT model achieves state-of-the-art (SOTA) micro-F1 performance (i.e., 0.940) on the \textsc{PubMed 20k RCT} dataset, its BIO tagging, which "blocks" different rhetorical sections in a paragraph, is not applicable to the multi-label setting.

On the single-labeled datasets, we find: (1) the LLM-based method delivers promising macro F1 results (\textsc{CS-abstract}: 0.716, \textsc{PubMed 20K RCT}: 0.879, \textsc{ART-CoreSC}: 0.282), demonstrating its effectiveness in balancing performance across classes. Unlike previous baseline methods that predict rhetorical labels based on each sentence embedding, \method\ leverages the contextual understand ability of LLM to grasp the whole context before generating the SSC label, therefore treating each class more equally; (2) the micro F1 scores reveal that \method's sample-specific performance is near SOTA (\textsc{CS-abstract}: 0.768, \textsc{PubMed 20K RCT}: 0.925, \textsc{ART-CoreSC}: 0.524) but does not outperform the SOTA, indicating a limitation in capturing the majority class compared to the fully fine-tuned baseline models.

Note that, different from the previous SOTA methods that tuned the parameters of the entire pre-trained model, \method\ is tuned using LoRA, keeping the original model parameters frozen while updates about 4\% additional parameters relative to the size of the entire LLM. 10,100,736 parameters are trainable, and 2,516,273,152 parameters are frozen. This approach significantly reduces storage requirements, as only the task-specific additional parameters need to be stored.

\subsubsection{Ablation Studies}

\begin{table}[]
\resizebox{\columnwidth}{!}{%
\begin{tabular}{l|cc|cc}
\hline
\multicolumn{1}{c|}{\multirow{2}{*}{Model}}                   & \multicolumn{2}{c|}{\textsc{biorc800}}                                 & \multicolumn{2}{c}{\textsc{CS-abstract}}                              \\ \cline{2-5} 
\multicolumn{1}{c|}{}                                         & \multicolumn{1}{l|}{Micro F1} & \multicolumn{1}{l|}{Macro F1} & \multicolumn{1}{l|}{Micro F1} & \multicolumn{1}{l}{Macro F1} \\ \hline
LLM-SSC                                                       & \multicolumn{1}{c|}{0.907}    & 0.912                         & \multicolumn{1}{c|}{0.768}    & 0.716                        \\ \hline
\begin{tabular}[c]{@{}l@{}}w/o \\ Demonstration\end{tabular}  & \multicolumn{1}{c|}{0.903}    & 0.911                         & \multicolumn{1}{c|}{0.742}    & 0.645                        \\ \hline
\begin{tabular}[c]{@{}l@{}}w/o \\ WeighCon\end{tabular}       & \multicolumn{1}{c|}{0.896}    & 0.901                         & \multicolumn{1}{c|}{0.746}    & 0.682                        \\ \hline
\begin{tabular}[c]{@{}l@{}}w/o \\ Space Thinking\end{tabular} & \multicolumn{1}{c|}{0.892}    & 0.899                         & \multicolumn{1}{c|}{0.749}    & 0.685                        \\ \hline
\end{tabular}%
}
\caption{Ablation study. }
\label{ablation_study}
\end{table}

We conduct ablation studies to assess the impact of various components of \method\ when testing on \textsc{biorc800} and \textsc{CS-abstract} as shown in Table~\ref{ablation_study}. Note that "w/o Space Thinking" refers to deleting space thinking mechanism by enabling the LLM to generate only one token directly after the prompt. For all four components, we observe the performance drops when each component is removed from \method, indicating that all four components in \method\ contribute to SSC performance on both single- and multi-label datasets. Note that the impact of each component is greater when the model is trained on \textsc{CS-Abstract} compared to \textsc{biorc800}. \textsc{CS-Abstract} consists of 654 abstracts with an average of 7.23 sentences per abstract, while \textsc{biorc800} contains 800 abstracts with an average of 9.89 sentences. The small size of the \textsc{CS-Abstract} dataset limits the model's performance, and adding three components mitigate the limitation. In contrast, this improvement is less evident on the larger \textsc{biorc800} dataset.


\subsubsection{Think before Speak Analysis}

\begin{table}[]
\resizebox{\columnwidth}{!}{%
\begin{tabular}{c|cc|cc}
\hline
\multirow{2}{*}{\begin{tabular}[c]{@{}c@{}}Number of\\ Generated Tokens\end{tabular}} &
  \multicolumn{2}{c|}{\textsc{biorc800}} &
  \multicolumn{2}{c}{\textsc{CS-abstract}} \\ \cline{2-5} 
 &
  \multicolumn{1}{l|}{Micro F1} &
  \multicolumn{1}{l|}{Macro F1} &
  \multicolumn{1}{l|}{Micro F1} &
  \multicolumn{1}{l}{Macro F1} \\ \hline
1 & \multicolumn{1}{c|}{0.897} & 0.904 & \multicolumn{1}{c|}{0.749} & 0.685 \\ \hline
2 & \multicolumn{1}{c|}{0.907} & 0.912 & \multicolumn{1}{c|}{0.768} & 0.716 \\ \hline
3 & \multicolumn{1}{c|}{0.895} & 0.904 & \multicolumn{1}{c|}{0.739} & 0.686 \\ \hline
\end{tabular}%
}
\caption{"Think before Speak" analysis results. Notice that generating the next one token equals to leaving no space for model to think. }
\label{Think_before_speak}
\end{table}

We analyze the "Think before Speak" mechanism to determine whether generating more tokens introduces more bias or provides space for model to "think". Table~\ref{Think_before_speak} presents the performance of the model when generating one, two, and three subsequent tokens. The results show that generating two tokens yields the best micro and macro F1 scores across both datasets. This suggests that generating two new tokens is sufficient to achieve optimal model performance for SSC task, whereas generating only one token restricts the model's ability to process information, and generating three tokens introduces bias into the SSC label prediction.

\section{Related Works}

\paragraph{SSC datasets}
SSC datasets are from various domains. \textsc{PubMed 20k RCT}~\cite{dernoncourt17} and \textsc{NICTA-PIBOSO}~\cite{kim2011automatic} are two datasets generally used in biomedical domain. \textsc{csabstruct}~\cite{ cohan-etal-2019-pretrained} and \textsc{cs-abstracts}~\cite{gonccalves2020deep} are datasets utilizing abstracts from computer science papers. \textsc{Emerald 100k}~\cite{stead2019emerald} and \textsc{MAZEA}~\cite{dayrell2012rhetorical} contains samples from multi-domains. In addition to these abstract-based datasets, some others use the full paper, such as \textsc{Dr. Inventor}~\cite{fisas2015discoursive} collecting samples from the computer graphics domain and \textsc{ART-CoreSC}~\cite{liakata-etal-2010-corpora} from physics, chemistry, and biochemistry domains. 

\paragraph{SSC methods} Before the deep learning paradigm, machine learning algorithms are applied to SSC~\cite{ruch2007using, mcknight2003categorization, lin2006generative}. These methods rely on hand-selected features and the classification performance is limited to the annotation amount and quality~\cite{brack2022cross}. Inspired by transfer learning and deep learning bringing pre-learned knowledge from external large datasets and simulates human-like thinking, recent SSC works leverage neural networks have been reported~\citep{agibetov2018fast,jin18,li21,shang2021span, brack2022cross, brack2024sequential}. The current SoTA methods commonly follow a hierarchical framework~\citep{brack2022cross}, including an encoding layer to represent word tokens (e.g. Word2Vec~\citep{mikolov2013distributed}) or embed sentences (e.g., CNN~\cite{albawi2017understanding}), followed by a context interaction layer to enrich the embedding using the surrounding context (e.g. Bi-LSTM~\cite{shang2021span}), and a labeling optimization layer to output the optimized sequential labels (e.g. CRF~\cite{yamada2020sequential}). In addition to the hierarchical framework, a BERT-based work leverages the BERT self-attention mechanism to handle the variable-length text by attending to features in context~\cite{cohan-etal-2019-pretrained}. 

\paragraph{Supervised Contrastive Learning with LLMs} 

Contrastive learning objectives could be widely applied in supervised LLM tasks. In text classification, these objectives enhance performance by providing a clearer understanding of class boundaries~\cite{chen2022contrastnet, pan2022improved, wang2022contrastive, zhang2022metadata, liao2024mask, zhang-etal-2024-generation, DBLP:conf/www/ZhengCHC24}. For named entity recognition, leveraging labeled entity types to create positive and negative pairs helps the model distinguish between different entities more effectively~\cite{das-etal-2022-container, huang2022copner, zhang2023reducing, zhang2022optimizing, mo2024mcl}. In semantic similarity evaluation, supervised contrastive learning improves the model's ability to recognize subtle semantic differences, thereby boosting task performance~\cite{gao-etal-2021-simcse, liang2024factorized}.

\section{Conclusion}
In this paper, we introduce \method, a unified framework for in-context learning and parameter-efficient LLM finetuning for multi-label sequential sentence classification problem. \method\ integrates complete contextual information within the prompt and considers neighboring context when making predictions. Additionally, we present a multi-label contrastive learning loss with auto-weighting scheme to reweight the importance of negative pairs and address the multi-label sequential sentence classification problem. Furthermore, we release \textsc{BIORC800}, a manually annotated multi-label SSC dataset using unstructured abstracts from the biomedical field, contributing to the development of more robust methodologies for this task. Extensive experiments validate the remarkable capability of \method\ in SSC tasks under both in-context learning and parameter-efficient finetuning settings.

\section{Limitations}

First, the \method\ requires a vast amounts of computational power and time to train due to the LLM size and complexity. For example, it takes over three hours to train a LLM-SSC model using \textsc{biorc800} on NVIDIA A100 GPU (5 epochs). Furthermore, the WeighCon mechanism introduced for multi-label SSC in this study could potentially be applied to other multi-label classification tasks; however, due to space limitations in this paper, a comprehensive exploration of the generalizability of this multi-label contrastive objective was not feasible. 

\section{Acknowledgement}

This work was partially supported by the National Library of Medicine of the National Institutes of Health under the award number R01LM014079. The content is solely the responsibility of the authors and does not necessarily represent the official views of the National Institutes of Health. The funder had no role in considering the study design or in the collection, analysis, interpretation of data, writing of the report, or decision to submit the article for publication.

\bibliography{custom}

\begin{thebibliography}{59}
\providecommand{\natexlab}[1]{#1}

\bibitem[{Agarwal and Yu(2009)}]{agarwal09}
Shashank Agarwal and Hong Yu. 2009.
\newblock Automatically classifying sentences in full-text biomedical articles into introduction, methods, results and discussion.
\newblock \emph{Bioinformatics}, 25(23):3174--3180.

\bibitem[{Agibetov et~al.(2018)Agibetov, Blagec, Xu, and Samwald}]{agibetov2018fast}
Asan Agibetov, Kathrin Blagec, Hong Xu, and Matthias Samwald. 2018.
\newblock Fast and scalable neural embedding models for biomedical sentence classification.
\newblock \emph{BMC bioinformatics}, 19(1):1--9.

\bibitem[{Albawi et~al.(2017)Albawi, Mohammed, and Al-Zawi}]{albawi2017understanding}
Saad Albawi, Tareq~Abed Mohammed, and Saad Al-Zawi. 2017.
\newblock Understanding of a convolutional neural network.
\newblock In \emph{2017 international conference on engineering and technology (ICET)}, pages 1--6. Ieee.

\bibitem[{Beltagy et~al.(2019)Beltagy, Lo, and Cohan}]{beltagy-etal-2019-scibert}
Iz~Beltagy, Kyle Lo, and Arman Cohan. 2019.
\newblock \href {https://doi.org/10.18653/v1/D19-1371} {{S}ci{BERT}: A pretrained language model for scientific text}.
\newblock In \emph{Proceedings of the 2019 Conference on Empirical Methods in Natural Language Processing and the 9th International Joint Conference on Natural Language Processing (EMNLP-IJCNLP)}, pages 3615--3620, Hong Kong, China. Association for Computational Linguistics.

\bibitem[{Brack et~al.(2024)Brack, Entrup, Stamatakis, Buscherm{\"o}hle, Hoppe, and Ewerth}]{brack2024sequential}
Arthur Brack, Elias Entrup, Markos Stamatakis, Pascal Buscherm{\"o}hle, Anett Hoppe, and Ralph Ewerth. 2024.
\newblock Sequential sentence classification in research papers using cross-domain multi-task learning.
\newblock \emph{International Journal on Digital Libraries (2024), online first}.

\bibitem[{Brack et~al.(2022)Brack, Hoppe, Buscherm{\"o}hle, and Ewerth}]{brack2022cross}
Arthur Brack, Anett Hoppe, Pascal Buscherm{\"o}hle, and Ralph Ewerth. 2022.
\newblock Cross-domain multi-task learning for sequential sentence classification in research papers.
\newblock In \emph{Proceedings of the 22nd ACM/IEEE Joint Conference on Digital Libraries}, pages 1--13.

\bibitem[{Chen et~al.(2022{\natexlab{a}})Chen, Zhang, Mao, and Xu}]{chen2022contrastnet}
Junfan Chen, Richong Zhang, Yongyi Mao, and Jie Xu. 2022{\natexlab{a}}.
\newblock Contrastnet: A contrastive learning framework for few-shot text classification.
\newblock In \emph{Proceedings of the AAAI Conference on Artificial Intelligence}, volume~36, pages 10492--10500.

\bibitem[{Chen et~al.(2022{\natexlab{b}})Chen, Zhang, Zheng, and Mao}]{chen2022dual}
Qianben Chen, Richong Zhang, Yaowei Zheng, and Yongyi Mao. 2022{\natexlab{b}}.
\newblock Dual contrastive learning: Text classification via label-aware data augmentation.
\newblock \emph{arXiv preprint arXiv:2201.08702}.

\bibitem[{Cohan et~al.(2019)Cohan, Beltagy, King, Dalvi, and Weld}]{cohan-etal-2019-pretrained}
Arman Cohan, Iz~Beltagy, Daniel King, Bhavana Dalvi, and Dan Weld. 2019.
\newblock \href {https://doi.org/10.18653/v1/D19-1383} {Pretrained language models for sequential sentence classification}.
\newblock In \emph{Proceedings of the 2019 Conference on Empirical Methods in Natural Language Processing and the 9th International Joint Conference on Natural Language Processing (EMNLP-IJCNLP)}, pages 3693--3699, Hong Kong, China. Association for Computational Linguistics.

\bibitem[{Das et~al.(2022)Das, Katiyar, Passonneau, and Zhang}]{das-etal-2022-container}
Sarkar Snigdha~Sarathi Das, Arzoo Katiyar, Rebecca Passonneau, and Rui Zhang. 2022.
\newblock \href {https://doi.org/10.18653/v1/2022.acl-long.439} {{CONT}ai{NER}: Few-shot named entity recognition via contrastive learning}.
\newblock In \emph{Proceedings of the 60th Annual Meeting of the Association for Computational Linguistics (Volume 1: Long Papers)}, pages 6338--6353, Dublin, Ireland. Association for Computational Linguistics.

\bibitem[{Dayrell et~al.(2012)Dayrell, Candido~Jr, Lima, Machado~Jr, Copestake, Feltrim, Tagnin, and Alu{\'\i}sio}]{dayrell2012rhetorical}
Carmen Dayrell, Arnaldo Candido~Jr, Gabriel Lima, Danilo Machado~Jr, Ann~A Copestake, Val{\'e}ria~Delisandra Feltrim, Stella~EO Tagnin, and Sandra~M Alu{\'\i}sio. 2012.
\newblock Rhetorical move detection in english abstracts: Multi-label sentence classifiers and their annotated corpora.
\newblock In \emph{LREC}, pages 1604--1609.

\bibitem[{Dernoncourt and Lee(2017)}]{dernoncourt2017pubmed}
Franck Dernoncourt and Ji~Young Lee. 2017.
\newblock Pubmed 200k {RCT:} a dataset for sequential sentence classification in medical abstracts.
\newblock In \emph{Proceedings of the Eighth International Joint Conference on Natural Language Processing, {IJCNLP} 2017, Taipei, Taiwan, November 27 - December 1, 2017, Volume 2: Short Papers}, pages 308--313. Asian Federation of Natural Language Processing.

\bibitem[{Dernoncourt et~al.(2017)Dernoncourt, Lee, and Szolovits}]{dernoncourt17}
Franck Dernoncourt, Ji~Young Lee, and Peter Szolovits. 2017.
\newblock \href {https://www.aclweb.org/anthology/E17-2110} {Neural networks for joint sentence classification in medical paper abstracts}.
\newblock In \emph{Proceedings of the 15th Conference of the {E}uropean Chapter of the Association for Computational Linguistics: Volume 2, Short Papers}, pages 694--700, Valencia, Spain. Association for Computational Linguistics.

\bibitem[{Devlin et~al.(2019)Devlin, Chang, Lee, and Toutanova}]{devlin2019bert}
Jacob Devlin, Ming-Wei Chang, Kenton Lee, and Kristina Toutanova. 2019.
\newblock Bert: Pre-training of deep bidirectional transformers for language understanding.
\newblock In \emph{Proceedings of the 2019 Conference of the North American Chapter of the Association for Computational Linguistics: Human Language Technologies, Volume 1 (Long and Short Papers)}, pages 4171--4186.

\bibitem[{Dong et~al.(2022)Dong, Li, Dai, Zheng, Wu, Chang, Sun, Xu, and Sui}]{dong2022survey}
Qingxiu Dong, Lei Li, Damai Dai, Ce~Zheng, Zhiyong Wu, Baobao Chang, Xu~Sun, Jingjing Xu, and Zhifang Sui. 2022.
\newblock A survey on in-context learning.
\newblock \emph{arXiv preprint arXiv:2301.00234}.

\bibitem[{Fisas et~al.(2015)Fisas, Saggion, and Ronzano}]{fisas2015discoursive}
Beatriz Fisas, Horacio Saggion, and Francesco Ronzano. 2015.
\newblock On the discoursive structure of computer graphics research papers.
\newblock In \emph{Proceedings of the 9th linguistic annotation workshop}, pages 42--51.

\bibitem[{Gao et~al.(2021)Gao, Yao, and Chen}]{gao-etal-2021-simcse}
Tianyu Gao, Xingcheng Yao, and Danqi Chen. 2021.
\newblock \href {https://doi.org/10.18653/v1/2021.emnlp-main.552} {{S}im{CSE}: Simple contrastive learning of sentence embeddings}.
\newblock In \emph{Proceedings of the 2021 Conference on Empirical Methods in Natural Language Processing}, pages 6894--6910, Online and Punta Cana, Dominican Republic. Association for Computational Linguistics.

\bibitem[{Gon{\c{c}}alves et~al.(2020)Gon{\c{c}}alves, Cortez, and Moro}]{gonccalves2020deep}
S{\'e}rgio Gon{\c{c}}alves, Paulo Cortez, and S{\'e}rgio Moro. 2020.
\newblock A deep learning classifier for sentence classification in biomedical and computer science abstracts.
\newblock \emph{Neural Computing and Applications}, 32:6793--6807.

\bibitem[{Goyal et~al.(2024)Goyal, Ji, Rawat, Menon, Kumar, and Nagarajan}]{goyal2023think}
Sachin Goyal, Ziwei Ji, Ankit~Singh Rawat, Aditya~Krishna Menon, Sanjiv Kumar, and Vaishnavh Nagarajan. 2024.
\newblock Think before you speak: Training language models with pause tokens.
\newblock In \emph{The Twelfth International Conference on Learning Representations, {ICLR} 2024, Vienna, Austria, May 7-11, 2024}.

\bibitem[{Gunel et~al.(2021)Gunel, Du, Conneau, and Stoyanov}]{DBLP:conf/iclr/GunelDCS21}
Beliz Gunel, Jingfei Du, Alexis Conneau, and Veselin Stoyanov. 2021.
\newblock Supervised contrastive learning for pre-trained language model fine-tuning.
\newblock In \emph{9th International Conference on Learning Representations, {ICLR} 2021, Virtual Event, Austria, May 3-7, 2021}. OpenReview.net.

\bibitem[{He et~al.(2020)He, Fan, Wu, Xie, and Girshick}]{he2020momentum}
Kaiming He, Haoqi Fan, Yuxin Wu, Saining Xie, and Ross Girshick. 2020.
\newblock Momentum contrast for unsupervised visual representation learning.
\newblock In \emph{Proceedings of the IEEE/CVF conference on computer vision and pattern recognition}, pages 9729--9738.

\bibitem[{Hu et~al.(2022)Hu, Shen, Wallis, Allen{-}Zhu, Li, Wang, Wang, and Chen}]{hu2021lora}
Edward~J. Hu, Yelong Shen, Phillip Wallis, Zeyuan Allen{-}Zhu, Yuanzhi Li, Shean Wang, Lu~Wang, and Weizhu Chen. 2022.
\newblock Lora: Low-rank adaptation of large language models.
\newblock In \emph{The Tenth International Conference on Learning Representations, {ICLR} 2022, Virtual Event, April 25-29, 2022}. OpenReview.net.

\bibitem[{Huang et~al.(2022)Huang, He, Wang, Zhang, Gong, Mao, and Li}]{huang2022copner}
Yucheng Huang, Kai He, Yige Wang, Xianli Zhang, Tieliang Gong, Rui Mao, and Chen Li. 2022.
\newblock Copner: Contrastive learning with prompt guiding for few-shot named entity recognition.
\newblock In \emph{Proceedings of the 29th International conference on computational linguistics}, pages 2515--2527.

\bibitem[{Jiang et~al.(2019)Jiang, Zhang, Ye, and Liu}]{jiang2019hierarchical}
Xinyu Jiang, Bowen Zhang, Yunming Ye, and Zhenhua Liu. 2019.
\newblock A hierarchical model with recurrent convolutional neural networks for sequential sentence classification.
\newblock In \emph{Natural Language Processing and Chinese Computing: 8th CCF International Conference, NLPCC 2019, Dunhuang, China, October 9--14, 2019, Proceedings, Part II 8}, pages 78--89. Springer.

\bibitem[{Jimeno~Yepes et~al.(2013)Jimeno~Yepes, Mork, and Aronson}]{yepes13}
Antonio Jimeno~Yepes, James Mork, and Alan Aronson. 2013.
\newblock \href {https://aclanthology.org/W13-1913} {Using the argumentative structure of scientific literature to improve information access}.
\newblock In \emph{Proceedings of the 2013 Workshop on Biomedical Natural Language Processing}, pages 102--110, Sofia, Bulgaria. Association for Computational Linguistics.

\bibitem[{Jin and Szolovits(2018)}]{jin18}
Di~Jin and Peter Szolovits. 2018.
\newblock \href {https://doi.org/10.18653/v1/D18-1349} {Hierarchical neural networks for sequential sentence classification in medical scientific abstracts}.
\newblock In \emph{Proceedings of the 2018 Conference on Empirical Methods in Natural Language Processing}, pages 3100--3109, Brussels, Belgium. Association for Computational Linguistics.

\bibitem[{Khosla et~al.(2020)Khosla, Teterwak, Wang, Sarna, Tian, Isola, Maschinot, Liu, and Krishnan}]{DBLP:conf/nips/KhoslaTWSTIMLK20}
Prannay Khosla, Piotr Teterwak, Chen Wang, Aaron Sarna, Yonglong Tian, Phillip Isola, Aaron Maschinot, Ce~Liu, and Dilip Krishnan. 2020.
\newblock Supervised contrastive learning.
\newblock In \emph{Advances in Neural Information Processing Systems 33: Annual Conference on Neural Information Processing Systems 2020, NeurIPS 2020, December 6-12, 2020, virtual}.

\bibitem[{Kim et~al.(2011)Kim, Martinez, Cavedon, and Yencken}]{kim2011automatic}
Su~Nam Kim, David Martinez, Lawrence Cavedon, and Lars Yencken. 2011.
\newblock Automatic classification of sentences to support evidence based medicine.
\newblock In \emph{BMC bioinformatics}, volume~12, pages 1--10. BioMed Central.

\bibitem[{Lan et~al.(2024)Lan, Cheng, Hoang, Ter~Riet, and Kilicoglu}]{lan2024automatic}
Mengfei Lan, Mandy Cheng, Linh Hoang, Gerben Ter~Riet, and Halil Kilicoglu. 2024.
\newblock Automatic categorization of self-acknowledged limitations in randomized controlled trial publications.
\newblock \emph{Journal of biomedical informatics}, 152:104628.

\bibitem[{Li et~al.(2021)Li, Burns, and Peng}]{li21}
Xiangci Li, Gully Burns, and Nanyun Peng. 2021.
\newblock Scientific discourse tagging for evidence extraction.
\newblock In \emph{Proceedings of the 16th Conference of the European Chapter of the Association for Computational Linguistics: Main Volume}, pages 2550--2562.

\bibitem[{Liakata et~al.(2010)Liakata, Teufel, Siddharthan, and Batchelor}]{liakata-etal-2010-corpora}
Maria Liakata, Simone Teufel, Advaith Siddharthan, and Colin Batchelor. 2010.
\newblock \href {http://www.lrec-conf.org/proceedings/lrec2010/pdf/644_Paper.pdf} {Corpora for the conceptualisation and zoning of scientific papers}.
\newblock In \emph{Proceedings of the Seventh International Conference on Language Resources and Evaluation ({LREC}'10)}, Valletta, Malta. European Language Resources Association (ELRA).

\bibitem[{Liang et~al.(2024)Liang, Deng, Ma, Zou, Morency, and Salakhutdinov}]{liang2024factorized}
Paul~Pu Liang, Zihao Deng, Martin~Q Ma, James~Y Zou, Louis-Philippe Morency, and Ruslan Salakhutdinov. 2024.
\newblock Factorized contrastive learning: Going beyond multi-view redundancy.
\newblock \emph{Advances in Neural Information Processing Systems}, 36.

\bibitem[{Liao et~al.(2024)Liao, Liu, Dai, Wu, Zhang, Huang, Chen, Jiang, Liu, Zhu et~al.}]{liao2024mask}
Wenxiong Liao, Zhengliang Liu, Haixing Dai, Zihao Wu, Yiyang Zhang, Xiaoke Huang, Yuzhong Chen, Xi~Jiang, David Liu, Dajiang Zhu, et~al. 2024.
\newblock Mask-guided bert for few-shot text classification.
\newblock \emph{Neurocomputing}, page 128576.

\bibitem[{Lin et~al.(2006)Lin, Karakos, Demner-Fushman, and Khudanpur}]{lin2006generative}
Jimmy Lin, Damianos Karakos, Dina Demner-Fushman, and Sanjeev Khudanpur. 2006.
\newblock Generative content models for structural analysis of medical abstracts.
\newblock In \emph{Proceedings of the hlt-naacl bionlp workshop on linking natural language and biology}, pages 65--72.

\bibitem[{McKnight and Srinivasan(2003)}]{mcknight2003categorization}
Larry McKnight and Padmini Srinivasan. 2003.
\newblock Categorization of sentence types in medical abstracts.
\newblock In \emph{AMIA annual symposium proceedings}, volume 2003, page 440. American Medical Informatics Association.

\bibitem[{Mikolov et~al.(2013)Mikolov, Sutskever, Chen, Corrado, and Dean}]{mikolov2013distributed}
Tomas Mikolov, Ilya Sutskever, Kai Chen, Greg~S Corrado, and Jeff Dean. 2013.
\newblock Distributed representations of words and phrases and their compositionality.
\newblock \emph{Advances in neural information processing systems}, 26.

\bibitem[{Mo et~al.(2024)Mo, Yang, Liu, Wang, Chen, Wang, and Li}]{mo2024mcl}
Ying Mo, Jian Yang, Jiahao Liu, Qifan Wang, Ruoyu Chen, Jingang Wang, and Zhoujun Li. 2024.
\newblock mcl-ner: Cross-lingual named entity recognition via multi-view contrastive learning.
\newblock In \emph{Proceedings of the AAAI Conference on Artificial Intelligence}, volume~38, pages 18789--18797.

\bibitem[{Moll{\'a}(2022)}]{molla2022overview}
Diego Moll{\'a}. 2022.
\newblock Overview of the 2022 alta shared task: Piboso sentence classification, 10 years later.
\newblock In \emph{Proceedings of the The 20th Annual Workshop of the Australasian Language Technology Association}, pages 178--182.

\bibitem[{Pan et~al.(2022)Pan, Hang, Sil, and Potdar}]{pan2022improved}
Lin Pan, Chung-Wei Hang, Avirup Sil, and Saloni Potdar. 2022.
\newblock Improved text classification via contrastive adversarial training.
\newblock In \emph{Proceedings of the AAAI Conference on Artificial Intelligence}, volume~36, pages 11130--11138.

\bibitem[{Pradhan et~al.(2003)Pradhan, Hacioglu, Ward, Martin, and Jurafsky}]{pradhan2003semantic}
Sameer Pradhan, Kadri Hacioglu, Wayne Ward, James~H Martin, and Daniel Jurafsky. 2003.
\newblock Semantic role parsing: Adding semantic structure to unstructured text.
\newblock In \emph{Third IEEE international conference on data mining}, pages 629--632. IEEE.

\bibitem[{Ruch et~al.(2007)Ruch, Boyer, Chichester, Tbahriti, Geissb{\"u}hler, Fabry, Gobeill, Pillet, Rebholz-Schuhmann, Lovis et~al.}]{ruch2007using}
Patrick Ruch, Celia Boyer, Christine Chichester, Imad Tbahriti, Antoine Geissb{\"u}hler, Paul Fabry, Julien Gobeill, Violaine Pillet, Dietrich Rebholz-Schuhmann, Christian Lovis, et~al. 2007.
\newblock Using argumentation to extract key sentences from biomedical abstracts.
\newblock \emph{International journal of medical informatics}, 76(2-3):195--200.

\bibitem[{Shang et~al.(2021)Shang, Ma, Lin, Yan, and Chen}]{shang2021span}
Xichen Shang, Qianli Ma, Zhenxi Lin, Jiangyue Yan, and Zipeng Chen. 2021.
\newblock A span-based dynamic local attention model for sequential sentence classification.
\newblock In \emph{Proceedings of the 59th Annual Meeting of the Association for Computational Linguistics and the 11th International Joint Conference on Natural Language Processing (Volume 2: Short Papers)}, pages 198--203.

\bibitem[{Stead et~al.(2019)Stead, Smith, Busch, and Vatanasakdakul}]{stead2019emerald}
Connor Stead, Stephen Smith, Peter Busch, and Savanid Vatanasakdakul. 2019.
\newblock Emerald 110k: a multidisciplinary dataset for abstract sentence classification.
\newblock In \emph{Proceedings of the The 17th Annual Workshop of the Australasian Language Technology Association}, pages 120--125.

\bibitem[{Sun et~al.(2023)Sun, Li, Li, Wu, Guo, Zhang, and Wang}]{sun-etal-2023-text}
Xiaofei Sun, Xiaoya Li, Jiwei Li, Fei Wu, Shangwei Guo, Tianwei Zhang, and Guoyin Wang. 2023.
\newblock \href {https://doi.org/10.18653/v1/2023.findings-emnlp.603} {Text classification via large language models}.
\newblock In \emph{Findings of the Association for Computational Linguistics: EMNLP 2023}, pages 8990--9005, Singapore. Association for Computational Linguistics.

\bibitem[{Team et~al.(2024)Team, Mesnard, Hardin, Dadashi, Bhupatiraju, Pathak, Sifre, Rivi{\`e}re, Kale, Love et~al.}]{team2024gemma}
Gemma Team, Thomas Mesnard, Cassidy Hardin, Robert Dadashi, Surya Bhupatiraju, Shreya Pathak, Laurent Sifre, Morgane Rivi{\`e}re, Mihir~Sanjay Kale, Juliette Love, et~al. 2024.
\newblock Gemma: Open models based on gemini research and technology.
\newblock \emph{arXiv preprint arXiv:2403.08295}.

\bibitem[{Teufel and Moens(1998)}]{teufel1998sentence}
Simone Teufel and Marc Moens. 1998.
\newblock Sentence extraction and rhetorical classification for flexible abstracts.
\newblock In \emph{AAAI Spring Symposium on Intelligent Text summarization}, pages 89--97.

\bibitem[{Wadhwa et~al.(2023)Wadhwa, Amir, and Wallace}]{wadhwa2023revisiting}
Somin Wadhwa, Silvio Amir, and Byron~C Wallace. 2023.
\newblock Revisiting relation extraction in the era of large language models.
\newblock In \emph{Proceedings of the conference. Association for Computational Linguistics. Meeting}, volume 2023, page 15566. NIH Public Access.

\bibitem[{Wang et~al.(2022)Wang, Dai et~al.}]{wang2022contrastive}
Ran Wang, Xinyu Dai, et~al. 2022.
\newblock Contrastive learning-enhanced nearest neighbor mechanism for multi-label text classification.
\newblock In \emph{Proceedings of the 60th Annual Meeting of the Association for Computational Linguistics (Volume 2: Short Papers)}, pages 672--679.

\bibitem[{Xie et~al.(2022)Xie, Hou, Yu, Zhang, Luo, and Zhu}]{xie2022multi}
Shaorong Xie, Chunning Hou, Hang Yu, Zhenyu Zhang, Xiangfeng Luo, and Nengjun Zhu. 2022.
\newblock Multi-label disaster text classification via supervised contrastive learning for social media data.
\newblock \emph{Computers and Electrical Engineering}, 104:108401.

\bibitem[{Yamada et~al.(2020)Yamada, Hirao, Sasano, Takeda, and Nagata}]{yamada2020sequential}
Kosuke Yamada, Tsutomu Hirao, Ryohei Sasano, Koichi Takeda, and Masaaki Nagata. 2020.
\newblock Sequential span classification with neural semi-markov crfs for biomedical abstracts.
\newblock In \emph{Findings of the Association for Computational Linguistics: EMNLP 2020}, pages 871--877.

\bibitem[{Zhang et~al.(2024)Zhang, Wang, and Yang}]{zhang-etal-2024-generation}
Ruohong Zhang, Yau-Shian Wang, and Yiming Yang. 2024.
\newblock \href {https://aclanthology.org/2024.eacl-long.39} {Generation-driven contrastive self-training for zero-shot text classification with instruction-following {LLM}}.
\newblock In \emph{Proceedings of the 18th Conference of the European Chapter of the Association for Computational Linguistics (Volume 1: Long Papers)}, pages 659--673, St. Julian{'}s, Malta. Association for Computational Linguistics.

\bibitem[{Zhang et~al.(2023{\natexlab{a}})Zhang, Cheng, Gao, and Poon}]{zhang2022optimizing}
Sheng Zhang, Hao Cheng, Jianfeng Gao, and Hoifung Poon. 2023{\natexlab{a}}.
\newblock Optimizing bi-encoder for named entity recognition via contrastive learning.
\newblock \emph{ICLR 2023}.

\bibitem[{Zhang et~al.(2023{\natexlab{b}})Zhang, Yuan, Li, and Liu}]{zhang2023reducing}
Xin Zhang, Jingling Yuan, Lin Li, and Jianquan Liu. 2023{\natexlab{b}}.
\newblock Reducing the bias of visual objects in multimodal named entity recognition.
\newblock In \emph{Proceedings of the Sixteenth ACM international conference on web search and data mining}, pages 958--966.

\bibitem[{Zhang et~al.(2022)Zhang, Shen, Wu, Xie, Hao, Wang, Wang, and Han}]{zhang2022metadata}
Yu~Zhang, Zhihong Shen, Chieh-Han Wu, Boya Xie, Junheng Hao, Ye-Yi Wang, Kuansan Wang, and Jiawei Han. 2022.
\newblock Metadata-induced contrastive learning for zero-shot multi-label text classification.
\newblock In \emph{Proceedings of the ACM Web Conference 2022}, pages 3162--3173.

\bibitem[{Zheng et~al.(2024{\natexlab{a}})Zheng, Chen, He, and Chen}]{DBLP:conf/www/ZhengCHC24}
Lecheng Zheng, Zhengzhang Chen, Jingrui He, and Haifeng Chen. 2024{\natexlab{a}}.
\newblock {MULAN:} multi-modal causal structure learning and root cause analysis for microservice systems.
\newblock In \emph{Proceedings of the {ACM} on Web Conference 2024, {WWW} 2024, Singapore, May 13-17, 2024}, pages 4107--4116. {ACM}.

\bibitem[{Zheng et~al.(2024{\natexlab{b}})Zheng, Jing, Li, Tong, and He}]{DBLP:conf/kdd/ZhengJLTH24}
Lecheng Zheng, Baoyu Jing, Zihao Li, Hanghang Tong, and Jingrui He. 2024{\natexlab{b}}.
\newblock Heterogeneous contrastive learning for foundation models and beyond.
\newblock In \emph{Proceedings of the 30th {ACM} {SIGKDD} Conference on Knowledge Discovery and Data Mining, {KDD} 2024, Barcelona, Spain, August 25-29, 2024}, pages 6666--6676. {ACM}.

\bibitem[{Zheng et~al.(2022)Zheng, Xiong, Zhu, and He}]{zheng2022contrastive}
Lecheng Zheng, Jinjun Xiong, Yada Zhu, and Jingrui He. 2022.
\newblock Contrastive learning with complex heterogeneity.
\newblock In \emph{Proceedings of the 28th ACM SIGKDD Conference on Knowledge Discovery and Data Mining}, pages 2594--2604.

\bibitem[{Zheng et~al.(2023)Zheng, Zhu, and He}]{DBLP:conf/sdm/ZhengZH23}
Lecheng Zheng, Yada Zhu, and Jingrui He. 2023.
\newblock Fairness-aware multi-view clustering.
\newblock In \emph{Proceedings of the 2023 {SIAM} International Conference on Data Mining, {SDM} 2023, Minneapolis-St. Paul Twin Cities, MN, USA, April 27-29, 2023}, pages 856--864. {SIAM}.

\bibitem[{Zheng et~al.(2021)Zheng, Wang, You, Qian, Zhang, Wang, and Xu}]{zheng2021weakly}
Mingkai Zheng, Fei Wang, Shan You, Chen Qian, Changshui Zhang, Xiaogang Wang, and Chang Xu. 2021.
\newblock Weakly supervised contrastive learning.
\newblock In \emph{Proceedings of the IEEE/CVF International Conference on Computer Vision}, pages 10042--10051.

\end{thebibliography}

\appendix

\section{In-context Learning / Task-specific Tuning Prompts}
\label{in_context_learning_prompts}

\begin{figure*}[p]
\centering
\includegraphics[width=16cm]{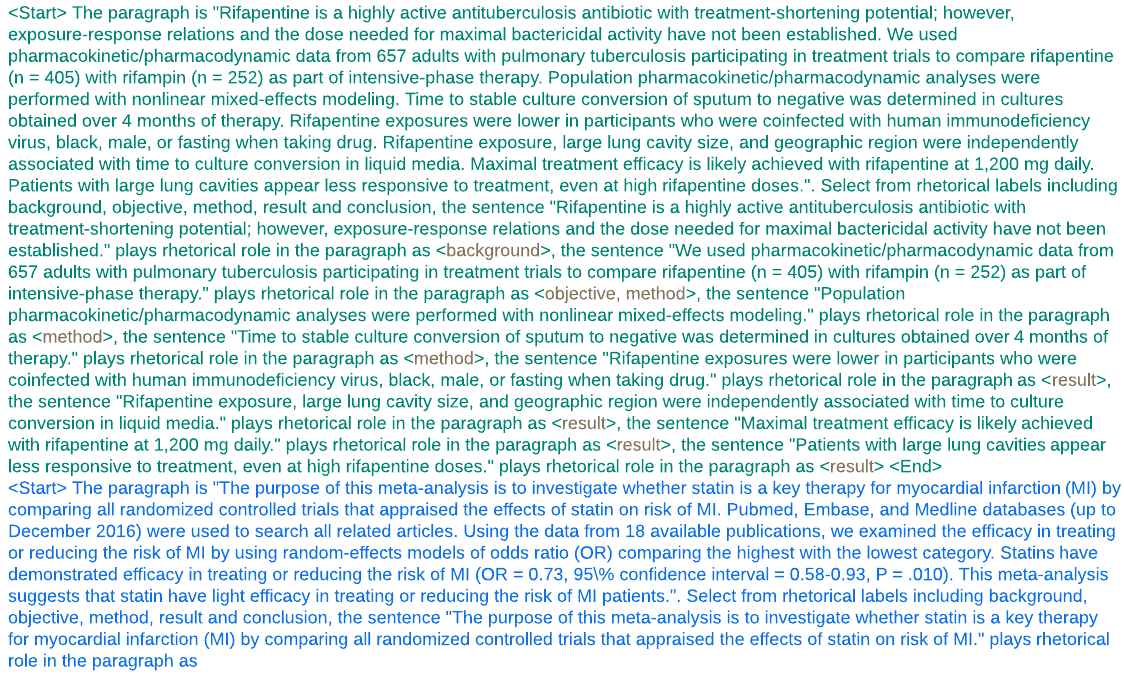}
\caption{\textsc{biorc800} 1-shot Prompt}
\label{biorc800_prompt}

\includegraphics[width=16cm]{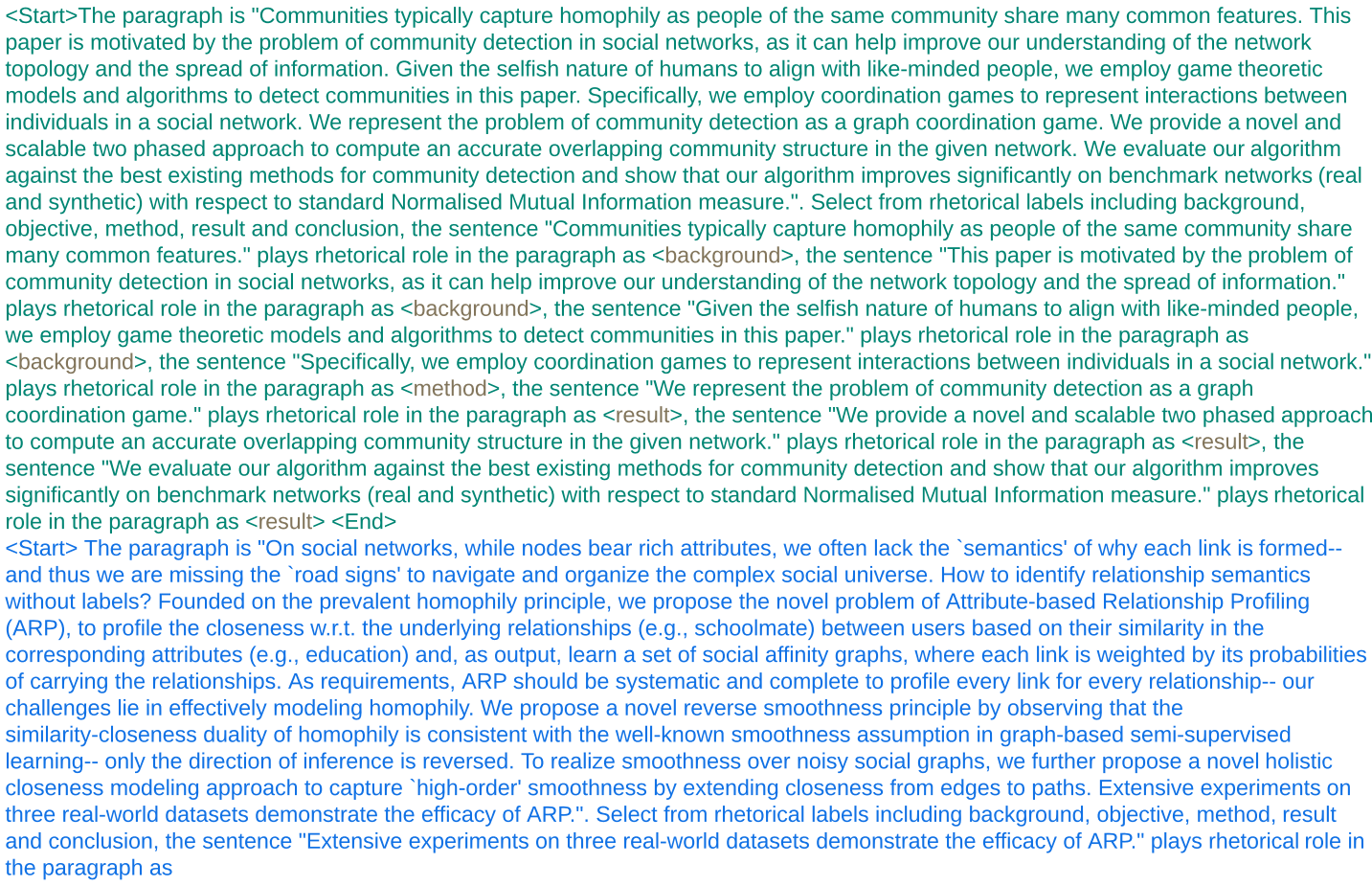}

\caption{\textsc{CS-Abstract} 1-shot Prompt}
\label{csabstract_prompt}

\end{figure*}

\begin{figure*}[p]
\centering
\includegraphics[width=16cm]{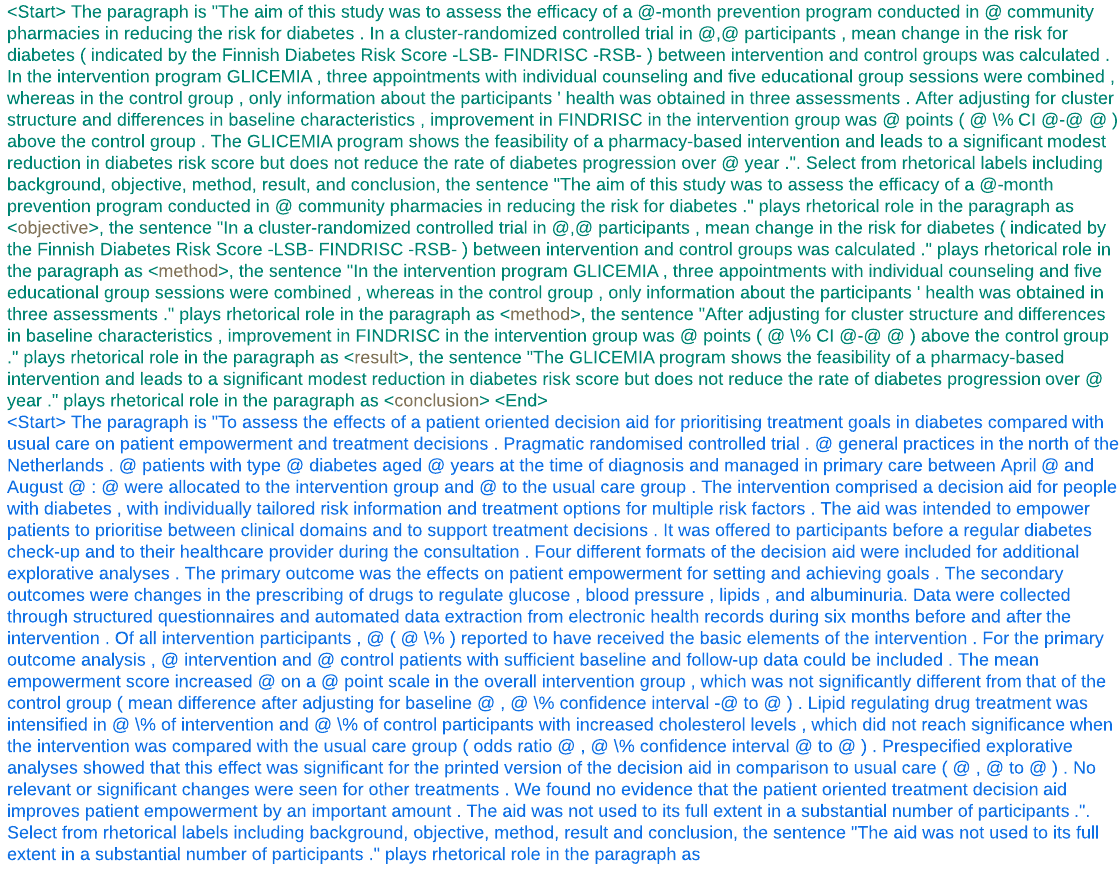}
\caption{\textsc{PubMed 20K RCT} 1-shot Prompt}
\label{pubmed_prompt}

\includegraphics[width=16cm]{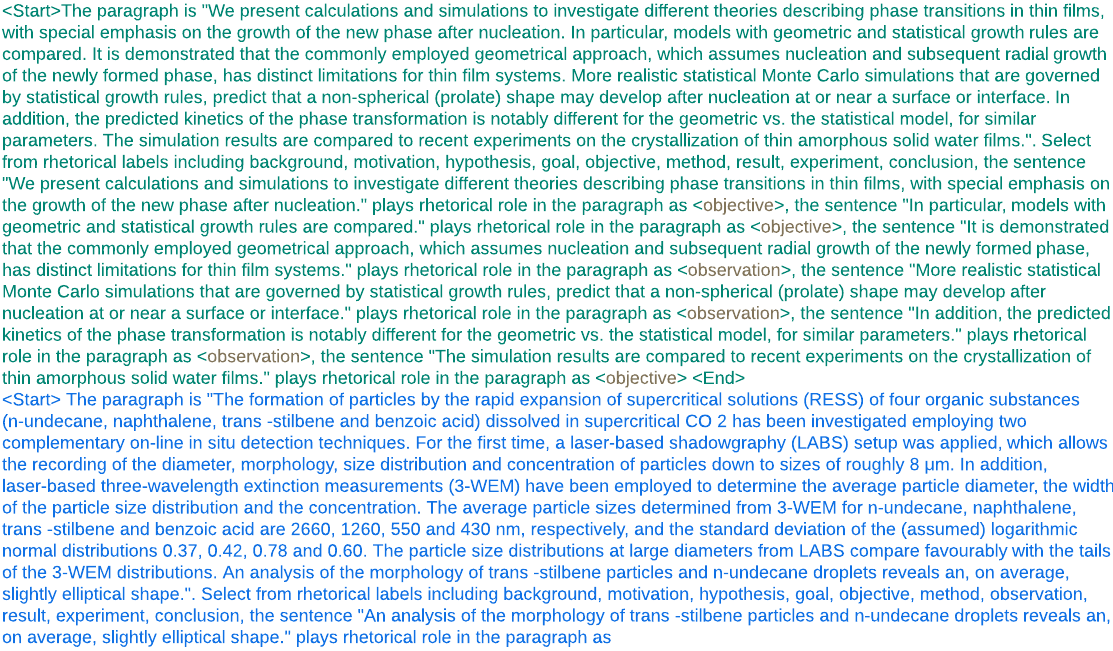}
\caption{\textsc{ART-CoreSC} 1-shot Prompt}
\label{art_coresc_prompt}

\end{figure*}

The one-shot in-context learning prompts, which are also utilized as task-specific model tuning instructions, are created for each dataset as figures~\ref{biorc800_prompt},~\ref{csabstract_prompt},~\ref{pubmed_prompt}, and ~\ref{art_coresc_prompt} show. In these figures, the demonstration portion of the prompt is highlighted in green, labels are marked in brown, and the query section is shown in blue.

\section{Annotation Guideline of \textsc{BIORC800}}
\label{guideline}
\paragraph{Introduction}

We aim to annotate PubMed abstracts for sequential rhetorical categories: BACKGROUND, OBJECTIVE, METHODS, RESULTS, and CONCLUSIONS (BOMRC schema).  Existing research on recognizing rhetorical categories from Randomized Controlled Trail (RCT) abstracts often use normalized labels derived from NLM mappings as ground truth label~\cite{jin18, cohan-etal-2019-pretrained, li21, brack2022cross}. This allows generating large amounts of training data for free, but our analysis shows that these labels are somewhat noisy. Additionally, though existing research have explored annotating the sequential sentences in other domain (e.g. computer science), no previous work attempted to extend the sequential sentence annotation from single label to multi-label. To address these issues, we will manually annotate 800 RCT abstracts using multi-label schema. The selected 800 RCT abstracts contain both structured (with NLM mapping; 100) and unstructured abstracts (without NLM mapping; 700). The structured abstracts with NLM mapping are involved to re-annotate since we hope to explore the noise introduced by auto-generation labeling. 

Each sentence needs to be annotated with one or more of the labels from [BACKGROUND, OBJECTIVE, METHODS, RESULTS, and CONCLUSIONS]. Insert all appropriate labels for a sentence in the “Annotated\_multi\_labels” column of the annotation spreadsheet and separate the labels by “,”. If a sentence is assigned with more than one labels, you should split the sentence by the rhetorical label and add split result in the column “Sentence\_split” following the format as: 

\textit{<OBJECTIVE>: We tested three models of this regularity <METHODS>: originally formulated for primate cerebral cortex, using quantitative data on the relative supragranular layer origins (SGN\%) of 151 projections from 19 areas (approximately 145,000 neurons) to four areas of cat extrastriate cortex}

Note that the sentences in an abstract are usually in the order of BACKGROUND – OBJECTIVE- METHODS- RESULTS- CONCLUSIONS, though some categories may be missing. For each abstract, they are not necessarily standardized to the 5 categories above. The Excel spreadsheet has the sentence ID for each sentence. To see the full abstract in context, you can click on the link in the third column, which will show the abstract in PubMed. If the abstract is structured, the “stru\_unstru” column value is “stru”, otherwise, “unstru”. Also note that some sentences do not fit into any of these categories. You can mark these as OTHER. For example, a sentence about study funding can be marked as OTHER.

Each annotator will be assigned 50 abstracts in the first round and another 50 abstracts in the second round. The abstracts assigned to each annotator in the first and second round will be the same, based on which to calculate the inter-annotator agreement. If the inter-annotator agreement (calculated by Cohen’s Kappa statistic) is lower than expected (0.85), a meeting will be scheduled for all annotators to discuss the cases that cause disagreement. The annotation created in the first and second rounds will be reconciled by one author to determine the golden standard. In the third round, the rest 700 abstracts will be evenly assigned to every annotators. Finally, the annotation results from the first round (50 abstracts), the second round (50), and the third round (225 abstracts from each of the four annotators) will be aggregated to the BIORC-1000 dataset. 

\paragraph{Points to Note}

There might be some situations that cause bias. Some points to follow to decrease the disagreements: 
\begin{itemize}
    \item With regards to the Objective/Background confusion, the presence of interrogative verbs, such as ‘evaluate’, ‘investigate’, ‘assess’, ‘aim’, or their derivational forms are strong clues for the Objective category. 
    \item Background sentences are often general statements about a disease, intervention, etc. (e.g., Diabetes mellitus is the common chronic metabolic disease’, ‘Regularity of laminar origin and termination of projections appears to be a common feature of corticocortical connections.’).

\item With regards to the Results/Conclusion confusion, we need to distinguish results obtained from the study (which often contain numbers, confidence intervals, etc.) from the implication/conclusion that can be drawn based on that result. One example we saw was a sentence about future work based on the results (‘The next step is external validation upon existence of independent trial data.’), which qualifies as Conclusion as it does not say anything about the results of this study. 

\item One outlier was an abstract about a disease guideline. I expect we may see some meta-analysis/review abstracts that may be similar. In this case, I annotated sentences summarizing the guidelines/recommendations as Conclusion sentences, and the sentence that describes what the guidelines are about as the Objective sentence. 

\item We try to annotate multi-label for each sentence. For example, if it is a sentence that sets the objective of the study but also mentions the method used, this should qualify as “OBJECTIVE, METHODS” sentence. (e.g., ‘We tested three models of this regularity, originally formulated for primate cerebral cortex, using quantitative data on the relative supragranular layer origins (SGN\%) of 151 projections from 19 areas to four areas of cat extrastriate cortex.’ is an “OBJECTIVE, METHODS” sentence). 

\item There may be some sentence splitting errors. Annotate such sentences, with the label of the actual sentence they would belong in. 

\item In reviews/surveys, the sentences that use the evidence from other articles should be annotated as RESULTS. A sample for review annotation (PMC26237111): 

\textit{Pressure ulcers (PUs) in individuals with spinal cord injury (SCI) present a persistent and costly problem. BACKGROUND}

\textit{Continuing effort in developing new technologies that support self-managed care is an important prevention strategy.     
 BACKGROUND}

\textit{Specifically, the aims of this scoping review are to review the key concepts and factors related to self-managed prevention of PUs in individuals with SCI and appraise the technologies available to assist patients in self-management of PU prevention practices. OBJECTIVE}

\textit{There is broad consensus that sustaining long-term adherence to prevention regimens is a major concern.    RESULTS} 

\textit{Recent literature highlights the interactions between behavioral and physiological risk factors.    RESULTS}

\textit{We identify four technology categories that support self-management: computer-based educational technologies demonstrated improved short-term gains in knowledge (2 studies), interface pressure mapping technologies demonstrated improved adherence to pressure-relief schedules up to 3 mo (5 studies), electrical stimulation confirmed improvements in tissue tolerance after 8 wk of training (3 studies), and telemedicine programs demonstrated improvements in independence and reduced hospital visits over 6 mo (2 studies). RESULTS }

...

However, if a sentence in the review abstract discusses a common finding shared by multiple previous papers or outlines the potential next steps for a topic, it should be annotated as “CONCLUSIONS”(PMC26237111):

\textit{Overall, self-management technologies demonstrated low-to-moderate effectiveness in addressing a subset of risk factors. CONCLUSIONS}

... 

\textit{However, the effectiveness of technologies in preventing PUs is limited due to a lack of incidence reporting. CONCLUSIONS}

\textit{In light of the key findings, we recommend developing integrated technologies that address multiple risk factors. CONCLUSIONS}

\item If the sentence starts with “we show that …” followed by some discussion of the experiment results, apply “OBJECTIVE, RESULTS”, or “RESULTS” to the sentence. Here is one example to illustrate the appropriate usage of “OBJECTIVE, RESULTS” (PMC23258531):

...

\textit{Aberrant expression of miR-31 has been found in various cancers, including colorectal cancer. BACKGROUND}

\textit{Here, we show that miR-31 is upregulated in human colon cancer tissues and cell lines, and that repression of miR-31 inhibited colon cancer cell proliferation and colony formation in soft agarose. OBJECTIVE, RESULTS}

\textit{To further elucidate the mechanism underlying the role of miR-31 in promoting colon cancer, we used online miRNA target prediction databases and found that the tumor suppressor RhoTBT1 may be a target of miR-31. METHODS}

…
\end{itemize}

\paragraph{Multilabel Samples}

Some of the multilabel annotations are presented as follows. Notice that these samples don't cover every type of multilabel case. Your annotations should adapt to the specific content of real cases, being more flexible as needed.

a.	OBJECTIVE-METHODS: 

\textit{We investigated this hypothesis using functional MRI and covariance analysis in 43 healthy skilled readers.}

\textit{<OBJECTIVE>: We investigated this hypothesis} 

\textit{<METHODS>: using functional MRI and covariance analysis in 43 healthy skilled readers.}

b.	RESULTS-CONCLUSION

\textit{The similarity of developmental features among the direct-developers suggests a correlation with mode of life history.}

\textit{<RESULTS>: The similarity of developmental features among the direct-developers}

\textit{<CONCLUSION>: suggests a correlation with mode of life history.}

c.	BACKGROUND-OBJECTIVE

\textit{Since the mPFC also projects heavily to NAc, we examined whether NAc-projecting pyramidal neurons also express 5-HT2A-R.}

\textit{<BACKGROUND>: Since the mPFC also projects heavily to NAc}

\textit{<OBJECTIVE>: we examined whether NAc-projecting pyramidal neurons also express 5-HT2A-R} 

d.	OBJECTIVE-METHODS

\textit{Here, we describe the features of MAP(2.0)3D server by analyzing, as an example, the cytochrome P450BM3 monooxygenase (CYP102A1).}

\textit{<OBJECTIVE>: Here, we describe the features of MAP(2.0)3D server}

\textit{<METHODS>: by analyzing, as an example, the cytochrome P450BM3 monooxygenase (CYP102A1)}

e.	Three-label

\textit{Although p38 MAPK activation usually promotes apoptosis, pharmacologic inhibition of p38 MAPK exacerbated OA-induced DNA fragmentation and loss of delta psi(m) in T leukemia cells, suggesting that, in this instance, the p38 MAPK signaling pathway promoted cell survival.}

\textit{<BACKGROUND>: p38 MAPK activation usually promotes apoptosis} 

\textit{<RESULTS>: pharmacologic inhibition of p38 MAPK exacerbated OA-induced DNA fragmentation and loss of delta psi(m) in T leukemia cells }

\textit{<CONCLUSIONS>: suggesting that, in this instance, the p38 MAPK signaling pathway promoted cell survival}

\section{\textsc{biorc800} Detailed Statistics}
\label{detailed_stat_append}
The detailed information of \textsc{biorc800} is shown in table~\ref{detailed_statistics}. 

\begin{table*}[]
\resizebox{\textwidth}{!}{%
\begin{tabular}{l|l|l|ll|l|l}
\hline
\multirow{2}{*}{} &
  \multirow{2}{*}{\begin{tabular}[c]{@{}l@{}}Number of \\ Structured Abstracts\end{tabular}} &
  \multirow{2}{*}{\begin{tabular}[c]{@{}l@{}}Number of \\ Unstructured Abstracts\end{tabular}} &
  \multicolumn{2}{l|}{Label Distribution} &
  \multirow{2}{*}{\begin{tabular}[c]{@{}l@{}}Average Sentences \\ per Abstracts\end{tabular}} &
  \multirow{2}{*}{\begin{tabular}[c]{@{}l@{}}Average Tokens per \\ Sentence/Abstract\end{tabular}} \\ \cline{4-5}
 &
   &
   &
  \multicolumn{1}{l|}{Label} &
  Distribution &
   &
   \\ \hline
Train &
  60 &
  420 &
  \multicolumn{1}{l|}{\begin{tabular}[c]{@{}l@{}}BACKGROUND\\ OBJECTIVE\\ METHODS\\ RESULTS\\ CONCLUSIONS\\ OTHER\end{tabular}} &
  \begin{tabular}[c]{@{}l@{}}805 (16.1\%)\\ 479 (9.6\%)\\ 1333 (26.6\%)\\ 1664 (33.2\%)\\ 654 (13.1\%)\\ 73 (1.5\%)\end{tabular} &
  9.86 &
  22.94/225.19 \\ \hline
Dev &
  20 &
  140 &
  \multicolumn{1}{l|}{\begin{tabular}[c]{@{}l@{}}BACKGROUND\\ OBJECTIVE\\ METHODS\\ RESULTS\\ CONCLUSIONS\\ OTHER\end{tabular}} &
  \begin{tabular}[c]{@{}l@{}}228 (13.2\%)\\ 184 (10.7\%)\\ 521 (30.3\%)\\ 528 (30.7\%)\\ 243 (14.1\%)\\ 17 (1.0\%)\end{tabular} &
  9.98 &
  22.97/229.26 \\ \hline
Test &
  20 &
  140 &
  \multicolumn{1}{l|}{\begin{tabular}[c]{@{}l@{}}BACKGROUND \\ OBJECTIVE\\ METHODS\\ RESULTS\\ CONCLUSIONS \\ OTHER\end{tabular}} &
  \begin{tabular}[c]{@{}l@{}}219 (13.3\%)\\ 164 (9.9\%)\\ 465 (28.1\%)\\ 565 (34.2\%)\\ 217 (13.1\%)\\ 22 (1.3\%)\end{tabular} &
  9.87 &
  23.43/231.21 \\ \hline
Total &
  100 &
  700 &
  \multicolumn{1}{l|}{\begin{tabular}[c]{@{}l@{}}BACKGROUND \\ OBJECTIVE\\ METHODS\\ RESULTS\\ CONCLUSIONS\\ OTHER\end{tabular}} &
  \begin{tabular}[c]{@{}l@{}}1252 (14.9\%)\\ 827 (9.9\%)\\ 2319 (27.7\%)\\ 2757 (32.9\%)\\ 1114 (13.3\%)\\ 112 (1.3\%)\end{tabular} &
  9.89 &
  23.03/227.80 \\ \hline
\end{tabular}%
}
\caption{\textsc{biorc800} Detailed Statistics}
\label{detailed_statistics}
\end{table*}
\section{Contextual Dependence Analysis}
\label{contextual_dependence}

We use two samples from \textsc{BIORC800} to illustrate the sequential sentence contextual dependencies.

\begin{itemize}

    \item PMID: 17448455

\textit{Based on this biological attribute we gain the possibility by means of using MSCs as the donors to develop a future cell therapy in clinical application.	BACKGROUND}

\textit{But using MSCs as donor cells inevitably raises \textbf{the question as to whether these donor cells would be immunogenic, and if so, would they be rejected after transplantation.	OBJECTIVE}}

\textit{\textbf{To investigate this}, human MSCs were cultured in vitro and induced to differentiate along neuronal lineage.	OBJECTIVE, METHODS}

\textit{The expression of human leukocyte antigen (HLA) class I and class II molecules and the co-stimulatory protein CD80 were increased on the surface of MSCs in the course of neuronal differentiation.	RESULTS}

\textit{But neither of the co-stimulatory proteins, CD40 or CD86, was expressed.	RESULTS} 

...

From the sample above, the OBJECTIVE of the work is indicated by the phrase "To investigate this" in the third sentence, with the pronoun "this" referring to "the question as to ..." mentioned in the previous sentence. Consequently, considering the context, the second sentence should be classified as OBJECTIVE rather than BACKGROUND.

\item PMID: 7231017

...

\textit{\textbf{Anserine supplementation increased superoxide dismutase (SOD) by 50\%} (p < 0.001, effect size d = 0.8 for both ANS-LD and ANS-HD), and preserved catalase (CAT) activity suggesting \textbf{an improved antioxidant activity}.	RESULTS, CONCLUSIONS}

...

\textit{There were slight but significant elevations in glutamate pyruvate transaminase (GPT) and creatine kinase isoenzyme (CKMB), especially in ANS-HD (p < 0.05) compared with ANS-LD or PLA.	RESULTS}

\textit{\textbf{Haematological biomarkers were largely unaffected by anserine}, its dose, and without interaction with post exercise time-course.	RESULTS}

\textit{However, compared with ANS-LD and PLA, ANS-HD increased the mean cell volume (MCV), and decreased the mean corpuscular haemoglobin concentration (MCHC) (p < 0.001).	RESULTS}

\textit{Anserine preserves cellular homoeostasis through enhanced antioxidant activity and protects cell integrity in healthy men, which is important for chronic disease prevention.	CONCLUSIONS}

\textit{However, \textbf{anserine temporal elevated exercise-induced cell-damage, together with enhanced antioxidant activity and haematological responses} suggest an augmented exercise-induced adaptative response and recovery.	RESULTS, CONCLUSIONS}

From the sample above, the sentence "However, anserine temporal elevated exercise..." should be labeled as "RESULTS, CONCLUSION" instead of "CONCLUSION" given the context of the whole paragraph. It is seen that "anserine temporal elevated exercise-induced cell-damage, together with enhanced antioxidant activity and haematological responses" is a summary of the previous RESULTS sentences and should be assigned with "RESULTS" label. 

\end{itemize}

\section{Semantic Coherence in Structured Abstracts Auto-generated Labels}
\label{semantic_coherence}
We start from two examples from PubMed 20K RCT to analyze the semantic coherence of the labels for structured abstracts. 

\begin{itemize}
    \item PMID: 25165090
        
    ...
    
    \textit{Although working smoke alarms halve deaths in residential fires , many households do not keep alarms operational.	BACKGROUND}

    \textit{\textbf{We tested whether theory-based education increases alarm operability}.	BACKGROUND}
    
    \textit{Randomised multiarm trial , with a single arm randomly selected for use each day , in low-income neighbourhoods in Maryland , USA .	METHODS}
    
    \textit{Intervention arms : ( 1 ) Full Education combining a health belief module with a social-cognitive theory module that provided hands-on practice installing alarm batteries and using the alarm 's hush button ; ( 2 ) Hands-on Practice social-cognitive module supplemented by typical fire department education ; ( 3 ) Current Norm receiving typical fire department education only.	METHODS}
    
    ...
    
    From the sample above, the sentence "We tested whether theory-based education increases alarm operability" should be labeled as "OBJECTIVES" instead of "BACKGROUND" given the context of the whole paragraph. 

\end{itemize}

\end{document}